\documentclass[runningheads]{llncs}

\usepackage{accv}

\usepackage{accvabbrv}

\usepackage{graphicx}
\usepackage{booktabs}
\usepackage{subcaption}

\usepackage[accsupp]{axessibility}  

\usepackage{hyperref}

\usepackage{orcidlink}
\usepackage[table]{xcolor}

\usepackage{multirow}
\begin{document}

\title{MonoLaDN: Language-Guided DeNoising for Monocular 3D Object Detection}

\titlerunning{MonoLaDN}

\author{Kiet Dang Vu\inst{1,2}\orcidlink{0009-0004-0964-8134} \and
Cao Le Minh Khoa\inst{1,2}\orcidlink{0009-0008-5056-6578} \and
Nguyen Viet Tri Pham\inst{1,2}\orcidlink{0009-0002-4205-0148} \and
An Nguyen Gia\inst{1,2}\orcidlink{0009-0004-7743-7013}\and
Tam Le Thuc Ho\inst{1,2}\orcidlink{0009-0008-5445-280X}\and
Duc Dung Nguyen\inst{1,2}\thanks{Corresponding author: \href{mailto:nddung@hcmut.edu.vn}{nddung@hcmut.edu.vn}}\orcidlink{0000-0001-7321-7401}}

\authorrunning{Kiet Dang Vu et al.}

\institute{AITech Lab., Ho Chi Minh City University of Technology (HCMUT) \and Vietnam National University Ho Chi Minh City (VNUHCM)}

\maketitle

\begin{abstract}
Monocular 3D object detection offers a low-cost route to 3D perception, but it remains fundamentally ill-posed, since depth must be inferred from a single view. Recent transformer-based detectors learn geometry purely from image features and 3D box annotations, lacking the high-level semantic priors that vision-language models (VLMs) acquire from large-scale pretraining. VLMs offer such priors, but deploying one directly as a detector is impractical due to its heavy inference cost. We propose MonoLaDN, a language-guided denoising framework that transfers the geometric reasoning of a VLM into a monocular 3D detector while keeping the deployed model purely vision-based. During training, a frozen VLM produces structured reasoning about each object's occlusion, depth, dimension, and angle from the image and its ground-truth boxes. We encode this reasoning into queries, perturb them with noise, and train the detector to reconstruct the clean reasoning. Reconstructing the corrupted reasoning forces the detector's depth-guided features to internalize the VLM's geometric priors, so this knowledge remains available even after the reasoning path is discarded at inference, adding no latency or parameters to the deployed model. Experiments on KITTI 3D and nuScenes show that MonoLaDN consistently outperforms state-of-the-art detectors and generalizes well to new domains.
\end{abstract}

\section{Introduction}
\label{sec:introduction}
\begin{figure}[h!]
    \centering
    \includegraphics[width=\textwidth]{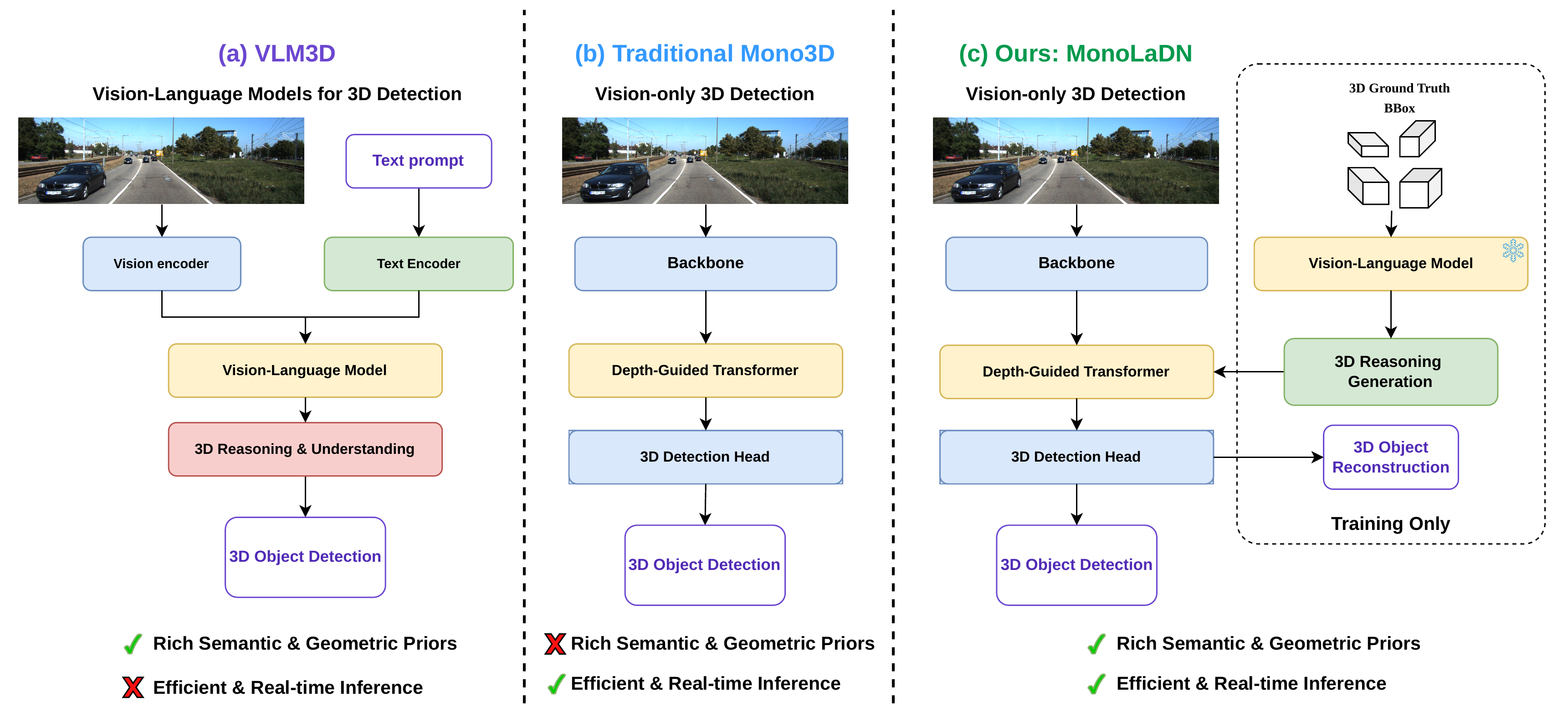}
   \caption{\textbf{Comparison of paradigms for monocular 3D detection.} (a) VLM-based methods carry rich semantic and geometric priors but are too costly for real-time inference. (b) Traditional vision-only detectors are efficient yet lack such priors, learning geometry only from image features and box annotations. (c) Our MonoLaDN injects VLM-derived 3D reasoning into a depth-guided detector during training, so it gains these priors while keeping a vision-only, efficient model at inference.}
    \label{fig:introduction:overview}
\end{figure}
Monocular 3D object detection aims to recover the location, dimension, and orientation of objects in 3D space from a single RGB image. Relying only on a standard camera, it avoids the cost, calibration burden, and hardware complexity of LiDAR~\cite{PeP,Centerbased3D,PV-RCNN} and multi-view camera~\cite{RayDN,BEVFormer,PETR} setups, offering a low-cost, easily deployable route to 3D perception for autonomous driving and robotics. The task is fundamentally ill-posed, however, since depth lost during projection must be inferred from a single view, where small distance errors translate into large 3D localization errors. These difficulties are compounded by occlusion and truncation, which remove the visual cues needed for reliable estimation.

Recent transformer-based detectors have substantially advanced the field~\cite{MonoDETR, MonoDGP, MonoCLUE, MonoDLGD}. MonoDETR~\cite{MonoDETR} introduced the first DETR~\cite{DETR} framework, guiding object queries with contextual depth cues rather than purely local features. MonoDGP~\cite{MonoDGP} improves geometry by replacing direct height regression with perspective-invariant geometry errors. MonoCLUE~\cite{MonoCLUE} strengthens appearance modeling via feature clustering and a generalized scene memory for robustness under occlusion, while MonoDLGD~\cite{MonoDLGD} improves supervision with a difficulty-aware label-guided denoising scheme. Despite this progress, these methods learn geometry purely from image features and 3D box annotations, without higher-level semantic knowledge about how objects are arranged and occluded in a scene.

In parallel, vision-language models (VLMs) have emerged as general-purpose perception systems with broad world knowledge and strong semantic reasoning. Their 3D competence, however, lags behind their 2D abilities, as benchmarks such as UniQA-3D~\cite{UniQA3D} show that state-of-the-art VLMs still trail specialized detectors on fundamental 3D tasks. A growing body of work therefore adapts VLMs toward explicit 3D understanding~\cite{SpatialReasoner, SpaceThinker, VLM3D, LocateAnything3D}, through explicit spatial reasoning~\cite{SpatialReasoner}, fine-tuning on synthetic reasoning traces~\cite{SpaceThinker}, end-to-end geometric perception~\cite{VLM3D}, or casting detection as next-token prediction~\cite{LocateAnything3D}. Yet these adapted models remain too large to run within a real-time detection pipeline, limiting their direct use as detectors.

These two lines of work occupy opposite extremes: VLMs encode rich semantic and geometric priors about real-world objects but are too computationally heavy for real-time deployment, whereas monocular detectors are efficient yet learn geometry solely from image features and box annotations, lacking such priors. The two are seldom reconciled, as a VLM's knowledge is presumed useful only when the VLM itself is retained at inference. This raises a natural question:
\begin{quote}
\itshape\bfseries Can the geometric reasoning of a VLM benefit a monocular 3D detector without incurring its inference cost?
\end{quote}
We answer affirmatively with MonoLaDN, a monocular 3D detector that exploits a VLM's geometric reasoning only during training and discards it entirely at inference. As illustrated in~\cref{fig:introduction:overview}, MonoLaDN bridges the two extremes: it keeps the lightweight, vision-only architecture of a depth-guided detector, while transferring the VLM's spatial knowledge into it through a training-only reasoning branch that is removed before deployment.

Concretely, during training, we pair each image with its ground-truth 3D boxes and feed them, through a prompt generator, to a frozen VLM that produces structured natural-language reasoning about each object's occlusion, depth, dimension, and angle. A text encoder maps this reasoning into attribute-specific embeddings, which are then perturbed with noise to form the reasoning queries. Inspired by the query denoising of DN-DETR~\cite{DN-DETR}, these noised queries are passed alongside the standard learnable detection queries into the depth-guided decoder, and the model is trained to reconstruct the clean object attributes from their corrupted reasoning. To prevent information leakage and control how reasoning is exchanged, we introduce a language-guided attention mask with two components. Following DN-DETR~\cite{DN-DETR}, a leakage-prevention mask blocks the learnable detection queries from attending to the reasoning queries, so they cannot copy the noised ground-truth as a shortcut that would collapse at inference. A separated-reasoning mask then restricts each reasoning query to three groups: all reasoning queries of the same object, those of the same attribute type across objects, and the learnable object queries, so each attribute is reconstructed from informative context without interference from unrelated attributes. At inference, the VLM, the reasoning branch, and the reasoning queries are all removed, leaving a vision-only detector that runs at the same speed as its MonoDGP~\cite{MonoDGP} baseline while benefiting from the knowledge absorbed during training.

\noindent Our contributions are summarized as follows:
\begin{itemize}
\item We propose MonoLaDN, the first framework to transfer the geometric reasoning of a vision-language model into a monocular 3D detector, using the VLM purely as training-time supervision and discarding it at inference, adding no latency or parameters to the deployed model.
\item We introduce a language-guided denoising scheme that encodes VLM's reasoning into attribute-specific queries and trains the detector to reconstruct the clean reasoning from their noised form, regulated by a language-guided attention mask that blocks ground-truth shortcuts and confines each attribute to its informative context.
\item Extensive experiments on KITTI 3D and nuScenes show that MonoLaDN consistently outperforms state-of-the-art monocular 3D detectors and generalizes to new domains, while adding no inference-time cost.
\end{itemize}
\section{Related Work}
\label{sec:related_work}
\textbf{3D - Vision Language Model.} Vision-language models such as Qwen2.5-VL~\cite{Qwen-VL} and InternVL~\cite{InternVL} excel at object recognition, 2D grounding, and visual reasoning. Their 3D competence lags behind, however, as Zuo et al.~\cite{UniQA3D} show through the UniQA-3D benchmark, where state-of-the-art VLMs perform poorly on fundamental 3D tasks. Recent work therefore adapts VLMs toward explicit 3D understanding: SpatialReasoner~\cite{SpatialReasoner} shares explicit 3D representations across perception, computation, and reasoning, while SpaceThinker~\cite{SpaceThinker} fine-tunes Qwen2.5-VL for quantitative spatial reasoning. Moving to geometric prediction, VLM-3D~\cite{VLM3D} enables end-to-end 3D perception via a joint semantic-geometric loss, and LocateAnything3D~\cite{LocateAnything3D} casts 3D detection as next-token prediction through a Chain-of-Sight sequence that grounds objects in 2D before inferring distance, size, and pose. Unlike these approaches, we use a VLM only during training to produce structured geometric reasoning about occlusion, depth, dimension, and angle, supervising a conventional monocular detector through language-guided denoising while keeping it VLM-free at inference.

\noindent\textbf{Monocular 3D Object Detection.} Transformer-based detectors now dominate monocular 3D object detection. MonoDETR~\cite{MonoDETR} pioneered this as the first DETR~\cite{DETR} framework for the task, using a depth-guided transformer whose decoder lets each object query draw features from depth-informative regions rather than only local cues. MonoDGP~\cite{MonoDGP} refines the geometric side, replacing direct height regression with perspective-invariant geometry errors and adding a decoupled visual decoder and a Region Segmentation Head. MonoCLUE~\cite{MonoCLUE} instead emphasizes appearance cues via K-means clustering and a generalized scene memory for robustness under occlusion, while MonoDLGD~\cite{MonoDLGD} addresses supervision with a difficulty-aware label-guided denoising scheme, built on DN-DETR~\cite{DN-DETR}, that perturbs and reconstructs ground-truth labels. Built on MonoDGP, our method applies language-guided denoising that, inspired by DN-DETR, perturbs and reconstructs VLM-derived geometric reasoning rather than ground-truth labels, learning geometry-aware representations grounded in the VLM's understanding while remaining VLM-free at inference.

\noindent\textbf{Vision-Language-to-Detector Distillation.} Distilling knowledge from pretrained VLM into detectors transfers their visual-semantic priors to lightweight models. ViLD~\cite{ViLD} established the paradigm, aligning a two-stage detector's box embeddings with a vision-language teacher's embeddings, and OADP~\cite{OADP} refined this with an object-aware distillation pyramid that preserves proposal features while transferring contextual knowledge. DK-DETR~\cite{DK-DETR} adapted distillation to query-based detectors via explicit semantic and implicit relational transfer. The idea has since reached 3D: CLIP-Fo3D~\cite{CLIP-Fo3D} distills dense 2D CLIP features into point clouds through projection, and ImOV3D~\cite{ImOV3D} transfers vision-language knowledge into a point-cloud detector by combining it with point-cloud geometry. In contrast to these methods, which transfer semantic and recognition priors into 2D or point-cloud detectors, we leverage a VLM's explicit geometric reasoning to guide a monocular 3D detector through language-guided denoising over occlusion, depth, dimension, and angle, adding no inference-time cost.
\section{Method}
\label{sec:method}
\subsection{Overview}
\label{sec:method:overview}
\vspace{-2em}
\begin{figure}[h!]
    \centering
    \includegraphics[width=\textwidth]{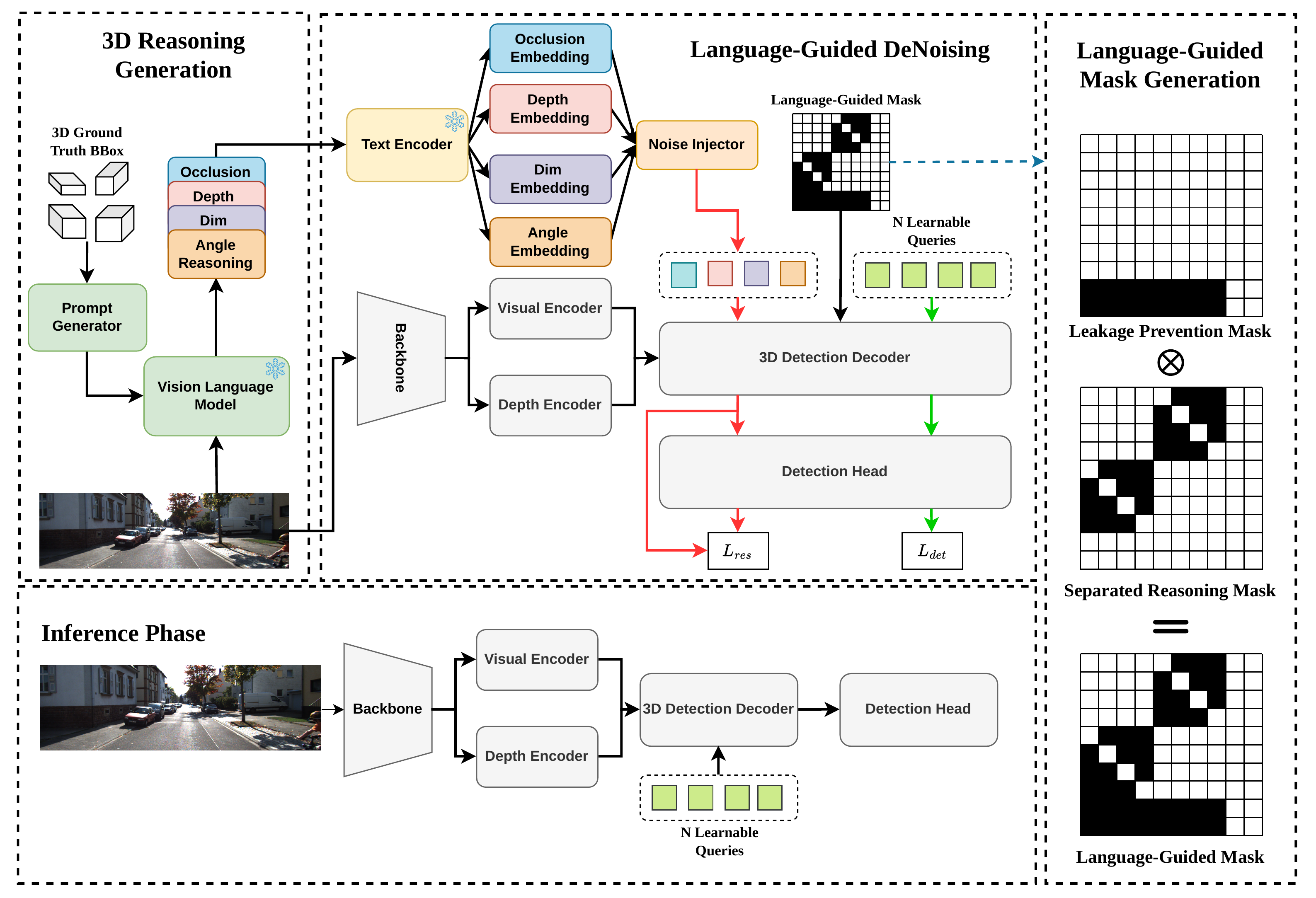}
    \caption{\textbf{Overview of MonoLaDN.} During training, a 3D Reasoning Generation stage queries a frozen VLM with the image and ground-truth 3D boxes to produce structured reasoning about each object's occlusion, depth, dimension, and angle. A text encoder and noise injector turn this reasoning into perturbed queries, decoded jointly with the $N$ learnable object queries under a Language-Guided Mask and trained to reconstruct the clean attributes, alongside the detection loss. At inference, the reasoning path is discarded, leaving a purely vision-based detector with no added cost.}
    \label{fig:method:overview:overall}
\end{figure}
\vspace{-1.5em}
\cref{fig:method:overview:overall} illustrates the overall architecture of MonoLaDN. The framework follows a dual-path design that shares a single depth-guided detector but differs between training and inference. We build on MonoDGP~\cite{MonoDGP} as the detection path, where a backbone extracts multi-scale features for a visual and a depth encoder, and $N$ learnable object queries are decoded into 3D boxes. This path is the only component retained at inference, incurring no cost over the baseline. Throughout, we index the four geometric attributes by $a \in \{occ, depth, dim, angle\}$ and denote by $M$ the number of ground-truth objects in an image.

During training, an auxiliary reasoning path supplies high-level supervision. A prompt generator pairs the image with its ground-truth 3D boxes and queries a frozen VLM to produce structured reasoning about each object's occlusion, depth, dimension, and angle (\cref{sec:method:3d_reasoning_gen}). A text encoder and noise injector then encode this reasoning into queries and perturb them with noise (\cref{sec:method:LaDN}). Decoded jointly with the learnable queries under a language-guided attention mask, these perturbed queries are trained to reconstruct the clean attributes via a reasoning loss $\mathcal{L}_{res}$ alongside the detection loss $\mathcal{L}_{det}$. This objective forces the detector's depth-guided features to internalize the VLM's reasoning, so the knowledge remains implicitly available once the reasoning path is discarded at inference.

\vspace{-1em}
\subsection{3D Reasoning Generation}
\label{sec:method:3d_reasoning_gen}
\vspace{-1em}
\begin{figure}[h!]
    \centering
    \includegraphics[width=\textwidth]{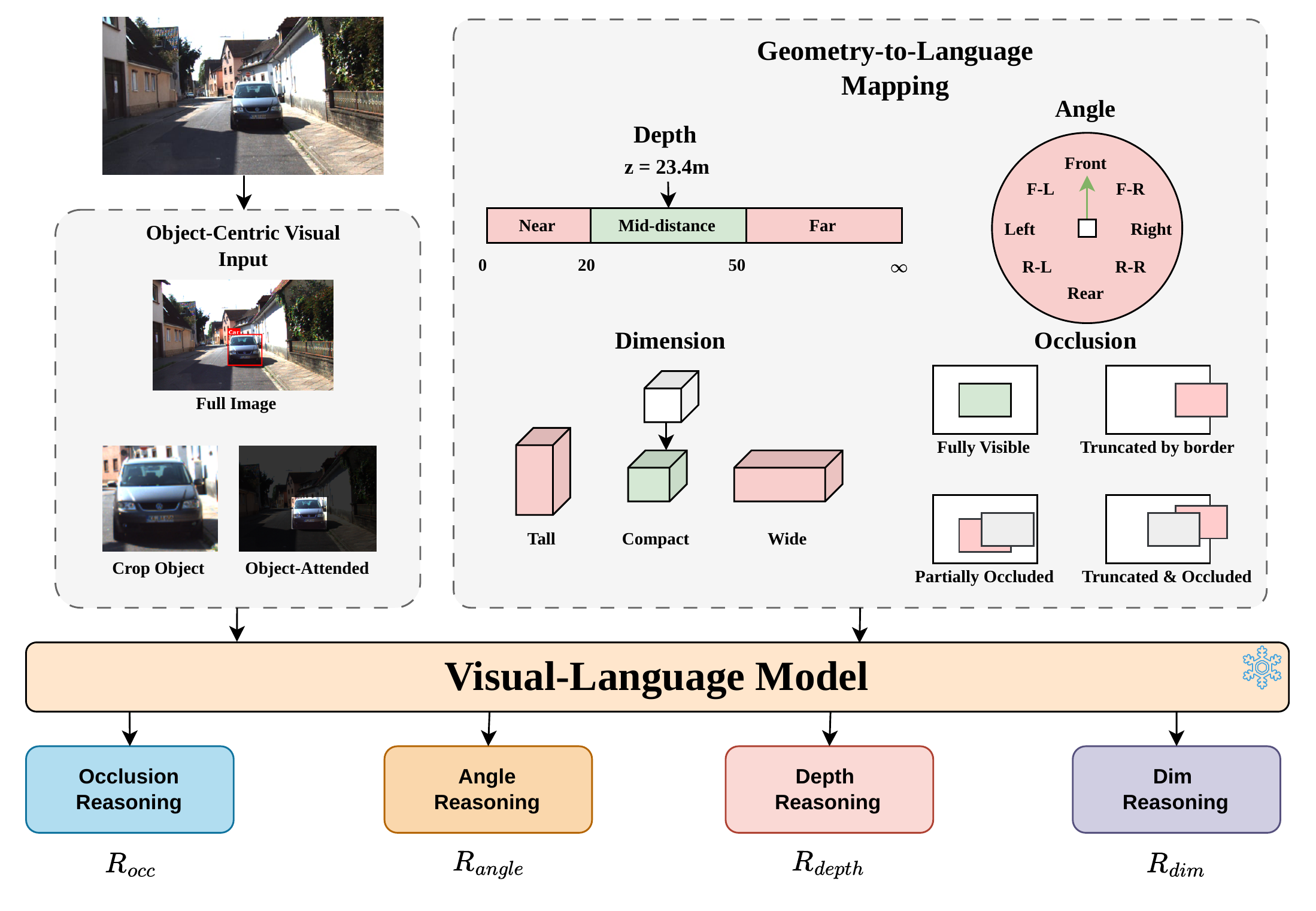}
    \caption{\textbf{3D Reasoning Generation.} The object-centric visual input and a geometry-to-language mapping are fed to a frozen VLM. The mapping discretizes each attribute into an ordinal label: depth by distance binning, angle by viewpoint sectoring, dimension by shape-archetype matching, and occlusion by combining truncation and occlusion flags. The VLM then produces structured reasoning per attribute.}
    \label{fig:method:overview:3d_reasoning_gen}
\end{figure}
For each ground-truth object, we generate structured language reasoning about four geometric attributes, namely occlusion, depth, dimension, and angle, which serves as the supervision signal for the denoising branch. As illustrated in \cref{fig:method:overview:3d_reasoning_gen}, this process comprises three stages: an object-centric visual input, a geometry-to-language mapping, and reasoning generation by a frozen VLM, all performed offline during training.

\noindent\textbf{Object-centric visual input.} Since a scene image typically contains multiple objects, prompting the VLM with the full image alone leaves the target ambiguous. For each object, we therefore construct three complementary views. The full image preserves global scene context such as road layout and inter-object overlap, the clean crop exposes the object's fine-grained appearance, and the object-attended view dims the background while preserving the original intensity of the object region. Together, these views focus the VLM on the target object while retaining the surrounding context required to reason about occlusion and depth.

\noindent\textbf{Geometry-to-language mapping.} As language models are known to handle discrete language far more reliably than precise continuous values~\cite{spithourakis2018numeracy, Sundararaman2020MethodsFN}, and this limitation persists for VLMs in 3D geometric estimation~\cite{UniQA3D}, we do not ask the VLM to regress geometry directly. Instead, we derive a discrete ordinal label (detailed in the supplementary material) for each attribute from the ground-truth 3D annotation: the occlusion and truncation flags are combined into a visibility label, the camera-frame depth is binned by distance, the box dimensions are matched to shape archetypes, and the observation angle is sectored into a viewpoint label. This procedure keeps the supervision grounded in the true geometry while expressing it in a form the VLM can reason about effectively.

\noindent\textbf{Reasoning generation.} Conditioned on the object-centric views and the discretized label, the frozen VLM produces one concise evidence sentence that justifies each attribute label from the visible cues. Pairing each label with its evidence yields the per-object reasoning set $\{R_{occ}, R_{depth}, R_{dim}, R_{angle}\}$, which is subsequently consumed by the Language-Guided DeNoising described in \cref{sec:method:LaDN}.

\subsection{Language-Guided DeNoising}
\label{sec:method:LaDN}
Given the object reasoning set $\{R_{occ}, R_{depth}, R_{dim}, R_{angle}\}$, we first convert it into a reasoning embedding set $\{E_{occ}, E_{depth}, E_{dim}, E_{angle}\}$ using a pretrained text encoder~\cite{CLIP,SigLIP}. To force the detector to internalize this reasoning rather than merely read it off at inference, we corrupt the reasoning embeddings with learned noise and train the model to reconstruct them, following the query-denoising paradigm of DN-DETR~\cite{DN-DETR}. Concretely, we inject noise by predicting a diagonal Gaussian $\mathcal{N}(\mu, \sigma^2)$ with a two-layer MLP and apply the reparameterization trick to obtain the perturbed query:
\begin{equation}
    q_{P} = \mu + \sigma \odot \epsilon, \qquad \epsilon \sim \mathcal{N}(0, I)
    \label{eq:reparam_trick}
\end{equation}
where $\mu$ and $\sigma$ are the predicted mean and standard deviation and $\odot$ is element-wise multiplication.

\noindent These perturbed queries are decoded jointly with the learnable object queries, so we must carefully design an attention mask to control how the two types of queries interact. Let $\mathbf{q}$ denote the set of $N$ learnable object queries, and let $\mathbf{q}_P$ denote the set of perturbed reasoning embeddings, where each of the $M$ ground-truth objects contributes four attribute-specific embeddings for occlusion, depth, dimension, and angle:
\begin{equation}
\begin{aligned}
\mathbf{q} &= \{q_0, q_1, \dots, q_{N-1}\}, \\
\mathbf{q}_P &= \{q_0^{occ}, q_0^{depth}, q_0^{dim}, q_0^{angle}, \dots, q_{M-1}^{occ}, q_{M-1}^{depth}, q_{M-1}^{dim}, q_{M-1}^{angle}\}.
\end{aligned}
\label{eq:query_set}
\end{equation}

\noindent We represent the attention mask as a binary matrix $A \in \{0,1\}^{W \times W}$ with $W = 4M + N$, where $A_{ij} = 1$ allows query $i$ to attend to query $j$ and $A_{ij} = 0$ blocks it. The $W$ queries are ordered as the $4M$ reasoning queries $\mathbf{q}_P$ followed by the $N$ learnable queries $\mathbf{q}$. We design this mask according to two rules, each addressing a different requirement, which we describe below before combining them into the final mask.

\noindent\textbf{Leakage Prevention Mask.} The reasoning queries are derived from the ground-truth 3D boxes. If the learnable queries could attend to them, the detector would simply copy this ground-truth information as a shortcut, and this dependency would break at inference, where the reasoning path is removed. Following DN-DETR~\cite{DN-DETR}, we therefore prevent every learnable query from attending to any reasoning query:
\begin{equation}
A^{lp}_{ij} =
\begin{cases}
0, & \text{if } i \in \mathbf{q} \text{ and } j \in \mathbf{q}_P, \\
1, & \text{otherwise,}
\end{cases}
\label{eq:leakage_mask}
\end{equation}
where $i \in \mathbf{q}$ denotes a learnable query and $j \in \mathbf{q}_P$ a reasoning query, so a learnable query cannot attend to any reasoning query while all other interactions remain open.

\noindent\textbf{Separated Reasoning Mask.} We further control how the reasoning queries exchange information, so that each attribute is reconstructed from related context rather than unrelated cues. Specifically, each reasoning query attends only to all reasoning queries of the same object, the reasoning queries of the same attribute across different objects, and the learnable object queries. Denoting a reasoning query by its object index $o$ and attribute $a \in \{occ, depth, dim, angle\}$, we define:
\begin{equation}
A^{sr}_{ij} =
\begin{cases}
0, & \text{if } i,j \in \mathbf{q}_P,\ o_i \neq o_j,\ \text{and}\ a_i \neq a_j, \\
1, & \text{otherwise,}
\end{cases}
\label{eq:separated_mask}
\end{equation}
so two reasoning queries are disconnected only when they belong to different objects and different attributes, while any interaction involving a learnable query stays open.

\noindent\textbf{Language-Guided Mask.} To enforce both rules at the same time, we combine the two masks by element-wise multiplication, so that an interaction is allowed only when both masks permit it:
\begin{equation}
A_{ij} = A^{lp}_{ij} \cdot A^{sr}_{ij}.
\label{eq:final_mask}
\end{equation}
\subsection{Loss Function}
\label{sec:method:loss_function}
\noindent\textbf{Detection Loss.} We adopt the detection objective of MonoDGP~\cite{MonoDGP}, which aggregates four terms: a region segmentation loss $\mathcal{L}_{\text{region}}$, a depth map regression loss $\mathcal{L}_{\text{depth}}$, a 2D detection loss $\mathcal{L}_{\text{2D}}$, and a 3D estimation loss $\mathcal{L}_{\text{3D}}$. The 2D term $\mathcal{L}_{\text{2D}}$ groups the classification, 2D box regression, GIoU, and projected-center objectives, whereas the 3D term $\mathcal{L}_{\text{3D}}$ supervises object size, angle, and center depth. The overall detection loss is given by
\begin{equation}
\mathcal{L}_{\text{det}} = \mathcal{L}_{\text{2D}} + \mathcal{L}_{\text{3D}} + \lambda \mathcal{L}_{\text{depth}} + \lambda \sum_{i=0}^{4} \mathcal{L}_{\text{region}}^{i},
\label{eq:detection_loss}
\end{equation}
where $\lambda$ weights the depth and region segmentation terms.

\noindent\textbf{Reconstruction Loss.} The reconstruction loss consists of two components. The first is an embedding reconstruction loss, which encourages the decoder output of each perturbed query to recover its original reasoning embedding produced by the text encoder, measured by a cosine similarity loss. Since the noise is sampled from a learnable diagonal Gaussian $\mathcal{N}(\mu, \sigma^2)$, we additionally regularize this distribution with a KL divergence term against a standard normal prior, following the variational autoencoder formulation~\cite{VAE}. This component is defined as
\begin{equation}
\mathcal{L}_{\text{emb}} = \sum_{a} \left( 1 - \cos\!\left(\hat{E}_a, E_a\right) \right) + \beta \, \mathrm{KL}\!\left(\mathcal{N}(\mu, \sigma^2) \,\|\, \mathcal{N}(0, I)\right),
\label{eq:embedding_loss}
\end{equation}
where $E_a$ is the reasoning embedding of attribute $a \in \{occ, depth, dim, angle\}$, $\hat{E}_a$ is the corresponding decoder output, and $\beta$ weights the KL term.

\begin{figure}[h!]
    \centering
    \includegraphics[width=0.9\columnwidth]{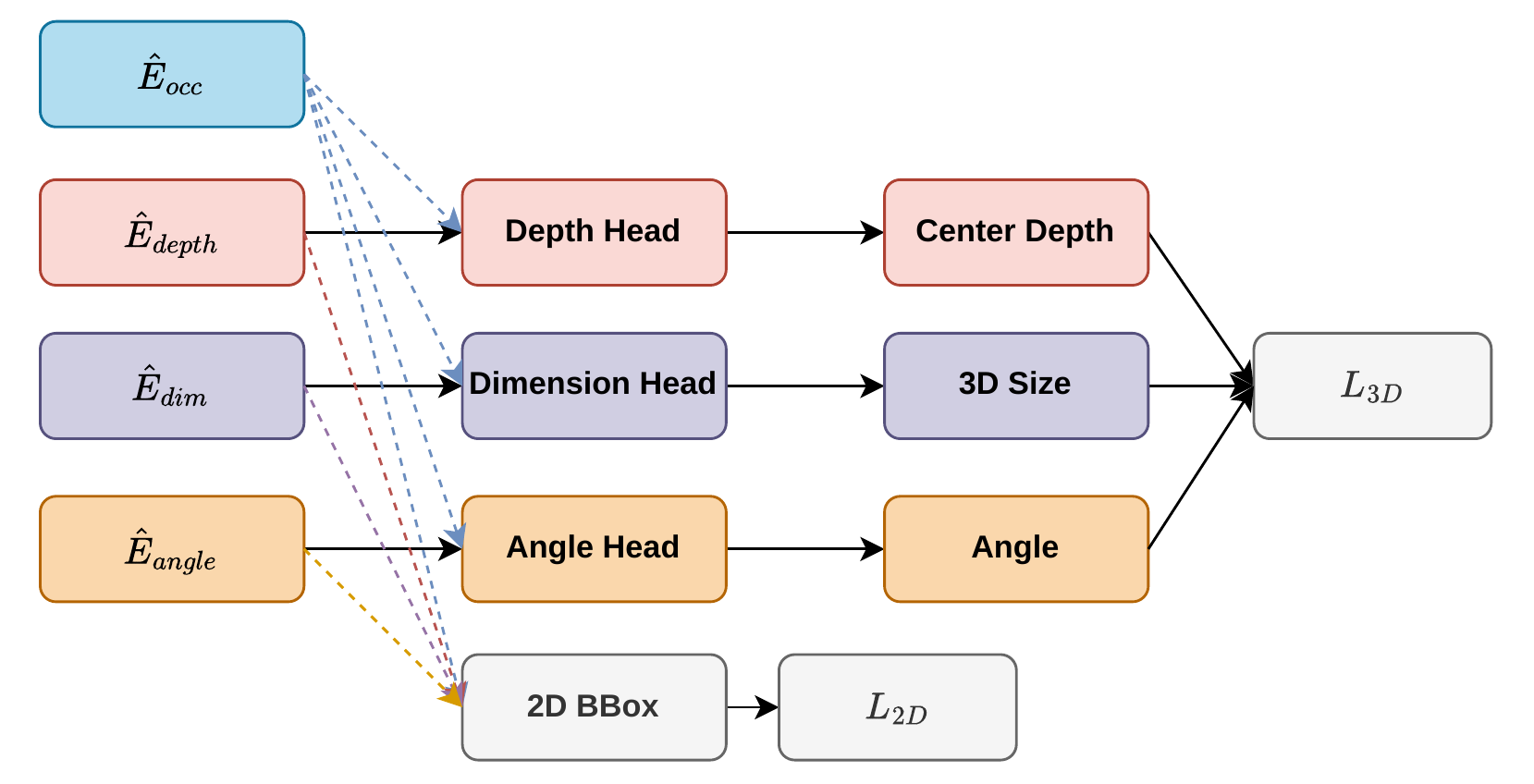}
    \caption{\textbf{Object reconstruction heads.} Each attribute output $\hat{E}_a$ is decoded into the box component it reasons about. The occlusion output $\hat{E}_{occ}$ has no box target of its own and is broadcast as shared visibility context (dashed) to the dimension, depth, and angle heads. When a head takes more than one embedding, the inputs are averaged before the head; the 2D center head uses the mean of all four outputs. The reconstructed components are supervised by the same 2D and 3D objectives as the detection loss.}
    \label{fig:method:obj_recon_heads}
\end{figure}

The second component is an object reconstruction loss, which requires the reasoning queries to reconstruct the 3D box of their object so that the learned representations carry genuine geometric meaning rather than merely matching the text embeddings. As illustrated in \cref{fig:method:obj_recon_heads}, each attribute output $\hat{E}_a$ is decoded into the box component it reasons about, aligning every reasoning query with a concrete geometric target and reinforcing attribute-specific specialization. The occlusion output $\hat{E}_{occ}$ has no box target of its own; since visibility governs how reliable the visual evidence is for the other attributes, it is instead broadcast as shared context to the dimension, depth, and angle heads. When a head receives more than one embedding, the inputs are averaged before being passed to the head, each attribute head takes the mean of its own output and $\hat{E}_{occ}$, and the 2D center head takes the mean of all four outputs $\bar{E} = \frac{1}{4}\sum_{a} \hat{E}_a$, since the projected center depends on the object as a whole rather than any single attribute. We supervise the resulting boxes with the same 2D and 3D objectives as the detection loss, giving $\mathcal{L}_{\text{obj}} = \mathcal{L}_{\text{2D}} + \mathcal{L}_{\text{3D}}$. The total reconstruction loss combines both components:
\begin{equation}
\mathcal{L}_{\text{res}} = \mathcal{L}_{\text{emb}} + \mathcal{L}_{\text{obj}}.
\label{eq:reconstruction_loss}
\end{equation}
\section{Experiments}
\label{sec:experiments}
\subsection{Settings}
\label{sec:experiments:setttings}
\textbf{Dataset.} We evaluated our model on the widely-used KITTI 3D object detection benchmark~\cite{KITTIDATA}, comprising 7481 training and 7518 testing images. Following Chen et al.~\cite{Chen20153DOP}, we further split the training data into 3712 training and 3769 validation images, allowing ablation studies on the contribution of each component in our MonoLaDN model.

\noindent\textbf{Evaluation metrics.} We assessed detection performance across three difficulty levels: easy, moderate, and hard. Two principal metrics were employed: $AP_{3D}$ and $AP_{BEV}$, which respectively quantify the accuracy of 3D bounding box predictions and the accuracy of their two-dimensional projections onto the bird's-eye view. Both metrics were computed at 40 recall positions~\cite{simonelli2019disentangling}.

\noindent\textbf{Implementation Details.}
We build MonoLaDN upon MonoDGP~\cite{MonoDGP} with a ResNet-50 backbone, keeping its detection loss $\mathcal{L}_{\text{det}}$ unchanged. The reasoning branch uses SpaceThinker-Qwen2.5-VL-3B~\cite{SpaceThinker} to generate the geometric reasoning text and SigLIP (base-patch16-224)~\cite{SigLIP} to embed it, and a $\beta$-VAE with fixed $\beta=4.0$ injects noise into these embeddings to form the perturbed queries. We train with AdamW for 250 epochs, using weight decay $10^{-4}$, an initial learning rate of $2\times10^{-4}$ halved at epochs 85, 125, 165, and 225, and batch size 8, on a single NVIDIA RTX 3090. At inference, the VLM and reasoning branch are removed, and predictions below 0.2 category confidence are filtered out.
\subsection{Main Results}
\label{sec:experiments:main_results}
\noindent\textbf{Experiment on the KITTI 3D test set.}
As shown in \cref{tab:main_result}, MonoLaDN establishes state-of-the-art performance on the KITTI test set for the car category. Without leveraging any auxiliary data, it attains 29.67, 20.45, and 18.02 $AP_{3D|R40}$ under the easy, moderate, and hard settings, surpassing all prior monocular approaches, including those using auxiliary LiDAR or depth supervision. Relative to our baseline MonoDGP~\cite{MonoDGP}, MonoLaDN yields improvements of +3.32, +1.73, and +2.05 $AP_{3D|R40}$ across the three difficulty levels, and it consistently exceeds the strongest competing methods, MonoDLGD~\cite{MonoDLGD} and MonoCLUE~\cite{MonoCLUE}, on all test metrics. The corresponding gains in $AP_{BEV|R40}$ further indicate that the improvement stems from more accurate 3D localization rather than image-plane alignment.

\noindent\textbf{Experiment on the KITTI 3D val set.}
A consistent trend is observed on the validation set, as presented in \cref{tab:main_result}. MonoLaDN achieves 35.21, 25.03, and 21.60 $AP_{3D|R40}$ under the easy, moderate, and hard settings, improving upon MonoDGP~\cite{MonoDGP} by +4.45, +2.69, and +2.58, respectively. It ranks first on the easy level and second on the moderate and hard levels, remaining highly competitive with MonoDLGD~\cite{MonoDLGD}. The consistent margins over the baseline on both $AP_{3D|R40}$ and $AP_{BEV|R40}$ demonstrate that transferring the geometric reasoning of the VLM during training yields more accurate 3D detection without incurring any inference cost.
\begin{table*}[ht!]
\centering
\caption{Comparisons with state-of-the-art monocular methods on the KITTI test and val sets for the car category. We \textbf{bold} the best results and \underline{underline} the second-best results. The {\color[HTML]{3166FF} blue} refers to the gain and the {\color[HTML]{FE0000} red} is the decrease of our method.}
\resizebox{\textwidth}{!}{%
\begin{tabular}{l|c|c|clcc|ccc|clcc|clcc}
\hline
\multicolumn{1}{c|}{} & & & \multicolumn{4}{c|}{Test, $AP_{3D|R40}$} & \multicolumn{3}{c|}{Test, $AP_{BEV|R40}$} & \multicolumn{4}{c|}{Val, $AP_{3D|R40}$} & \multicolumn{4}{c}{Val, $AP_{BEV|R40}$} \\ \cline{4-18}
\multicolumn{1}{l|}{\multirow{-2}{*}{Methods}} & \multirow{-2}{*}{Extra data} & \multirow{-2}{*}{Venue} & \multicolumn{2}{c}{Easy} & Mod. & Hard & Easy & Mod. & Hard & \multicolumn{2}{c}{Easy} & Mod. & Hard & \multicolumn{2}{c}{Easy} & Mod. & Hard \\ \hline
MonoDTR~\cite{MonoDTR} & & CVPR 2022 & \multicolumn{2}{c}{21.99} & 15.39 & 12.73 & 28.59 & 20.38 & 17.14 & \multicolumn{2}{c}{24.52} & 18.57 & 15.51 & \multicolumn{2}{c}{33.33} & 25.35 & 21.68 \\
DID-M3D~\cite{DID-M3D} &  & ECCV 2022 & \multicolumn{2}{c}{24.40} & 16.29 & 13.75 & 32.95 & 22.76 & 19.83 & \multicolumn{2}{c}{22.98} & 16.12 & 14.03 & \multicolumn{2}{c}{31.10} & 22.76 & 19.50 \\
OccupancyM3D~\cite{OccupanyM3D} & \multirow{-3}{*}{LiDAR} & CVPR 2024 & \multicolumn{2}{c}{25.55} & 17.02 & 14.79 & 35.38 & 24.18 & 21.37 & \multicolumn{2}{c}{26.87} & 19.96 & 17.15 & \multicolumn{2}{c}{35.72} & 26.60 & 23.68 \\ \hline
MonoPGC~\cite{MonoPGC} &  & ICRA 2023 & \multicolumn{2}{c}{24.68} & 17.17 & 14.14 & 32.50 & 23.14 & 20.30 & \multicolumn{2}{c}{25.67} & 18.63 & 15.65 & \multicolumn{2}{c}{34.06} & 24.26 & 20.78 \\
OPA-3D~\cite{OPA-3D} & \multirow{-2}{*}{Depth} & RAL 2023 & \multicolumn{2}{c}{24.60} & 17.05 & 14.25 & 33.54 & 22.53 & 19.22 & \multicolumn{2}{c}{24.97} & 19.40 & 16.59 & \multicolumn{2}{c}{33.80} & 25.51 & 22.13 \\ \hline
MonoCon~\cite{MonoCon} &  & AAAI 2022 & \multicolumn{2}{c}{22.50} & 16.46 & 13.95 & 31.12 & 22.10 & 19.00 & \multicolumn{2}{c}{26.33} & 19.01 & 15.98 & \multicolumn{2}{c}{-} & - & - \\
DEVIANT~\cite{DEVIANT} &  & ECCV 2022 & \multicolumn{2}{c}{21.88} & 14.46 & 11.89 & 29.65 & 20.44 & 17.43 & \multicolumn{2}{c}{24.63} & 16.54 & 14.52 & \multicolumn{2}{c}{32.60} & 23.04 & 19.99 \\
MonoDDE~\cite{MonoDDE} &  & CVPR 2022 & \multicolumn{2}{c}{24.93} & 17.14 & 15.10 & 33.58 & 23.46 & 20.37 & \multicolumn{2}{c}{26.66} & 19.75 & 16.72 & \multicolumn{2}{c}{35.51} & 26.48 & 23.07 \\
MonoUNI~\cite{MonoUNI} &  & NeurIPS 2023 & \multicolumn{2}{c}{24.75} & 16.73 & 13.49 & - & - & - & \multicolumn{2}{c}{24.51} & 17.18 & 14.01 & \multicolumn{2}{c}{-} & - & - \\
MonoDETR~\cite{MonoDETR} &  & ICCV 2023 & \multicolumn{2}{c}{25.00} & 16.47 & 13.58 & 33.60 & 22.11 & 18.60 & \multicolumn{2}{c}{28.84} & 20.61 & 16.38 & \multicolumn{2}{c}{37.86} & 26.95 & 22.80 \\
MonoCD~\cite{MonoCD} &  & CVPR 2024 & \multicolumn{2}{c}{25.53} & 16.59 & 14.53 & 33.41 & 22.81 & 19.57 & \multicolumn{2}{c}{26.45} & 19.37 & 16.38 & \multicolumn{2}{c}{34.60} & 24.96 & 21.51 \\
FD3D~\cite{FD3D} & & AAAI 2024 & \multicolumn{2}{c}{25.38} & 17.12 & 14.50 & 34.20 & 23.72 & 20.76 & \multicolumn{2}{c}{28.22} & 20.23 & 17.04 & \multicolumn{2}{c}{36.98} & 26.77 & 23.16 \\
MonoDGP~\cite{MonoDGP} & & CVPR 2025 & \multicolumn{2}{c}{26.35} & 18.72 & 15.97 & 35.24 & 25.23 & 22.02 & \multicolumn{2}{c}{30.76} & 22.34 & 19.02 & \multicolumn{2}{c}{39.40} & 28.20 & 24.42 \\
MonoDLGD~\cite{MonoDLGD} & & AAAI 2026 & \multicolumn{2}{c}{\underline{29.11}} & \underline{19.87} & \underline{17.74} & \underline{36.63} & 25.30 & \underline{23.13} & \multicolumn{2}{c}{\underline{34.89}} & \textbf{25.19} & \textbf{21.78} & \multicolumn{2}{c}{41.68} & \underline{30.53} & \underline{27.76} \\
MonoCLUE~\cite{MonoCLUE} & \multirow{-10}{*}{None} & AAAI 2026 & \multicolumn{2}{c}{27.94} & 19.70 & 16.69 & 36.15 & \underline{26.15} & 22.81 & \multicolumn{2}{c}{33.74} & 24.10 & 20.58 & \multicolumn{2}{c}{\underline{41.79}} & 29.91 & 26.00 \\ \hline
MonoLaDN (Ours) & None & - & \multicolumn{2}{c}{\textbf{29.67}} & \textbf{20.45} & \textbf{18.02} & \textbf{36.94} & \textbf{26.81} & \textbf{23.73} & \multicolumn{2}{c}{\textbf{35.21}} & \underline{25.03} & \underline{21.60} & \multicolumn{2}{c}{\textbf{42.09}} & \textbf{30.75} & \textbf{27.89} \\
Improvement vs second-best & - & - & \multicolumn{2}{c}{{\color[HTML]{3166FF}+0.56}} & {\color[HTML]{3166FF}+0.58} & {\color[HTML]{3166FF}+0.28} & {\color[HTML]{3166FF}+0.31} & {\color[HTML]{3166FF}+0.66} & {\color[HTML]{3166FF}+0.60} & \multicolumn{2}{c}{{\color[HTML]{3166FF}+0.32}} & {\color[HTML]{FE0000}-0.16} & {\color[HTML]{FE0000}-0.18} & \multicolumn{2}{c}{{\color[HTML]{3166FF}+0.30}} & {\color[HTML]{3166FF}+0.22} & {\color[HTML]{3166FF}+0.13} \\
Improvement vs MonoDGP & - & - & \multicolumn{2}{c}{{\color[HTML]{3166FF}+3.32}} & {\color[HTML]{3166FF}+1.73} & {\color[HTML]{3166FF}+2.05} & {\color[HTML]{3166FF}+1.70} & {\color[HTML]{3166FF}+1.58} & {\color[HTML]{3166FF}+1.71} & \multicolumn{2}{c}{{\color[HTML]{3166FF}+4.45}} & {\color[HTML]{3166FF}+2.69} & {\color[HTML]{3166FF}+2.58} & \multicolumn{2}{c}{{\color[HTML]{3166FF}+2.69}} & {\color[HTML]{3166FF}+2.55} & {\color[HTML]{3166FF}+3.47} \\ \hline
\end{tabular}}
\label{tab:main_result}
\end{table*}

\begin{table*}[t]
\centering
\caption{\textbf{Cross-dataset depth accuracy and computational complexity.} (a) Depth prediction MAE ($\downarrow$) on KITTI and nuScenes frontal validation. (b) Complexity on a single RTX 3090 GPU with batch size 1. \textbf{Best} in bold, \underline{second best} underlined.}
\label{tab:cross_complexity}
\begin{subtable}{0.50\textwidth}
\centering
\caption{Cross-dataset depth prediction MAE ($\downarrow$).}
\setlength{\tabcolsep}{5pt}
\renewcommand{\arraystretch}{1.15}
\resizebox{\linewidth}{!}{%
\begin{tabular}{l|cccc|cccc}
\toprule
\multicolumn{1}{c|}{\multirow{2}{*}{Method}} & \multicolumn{4}{c|}{KITTI Val} & \multicolumn{4}{c}{nuScenes frontal Val} \\
\multicolumn{1}{c|}{} & 0--20 & 20--40 & 40--$\infty$ & All & 0--20 & 20--40 & 40--$\infty$ & All \\
\midrule
GUPNet~\cite{GUPNet}     & 0.45 & 1.10 & 1.85 & 0.89 & 0.82 & 1.70 & 6.20 & 1.45 \\
DEVIANT~\cite{DEVIANT}   & 0.40 & 1.09 & 1.80 & 0.87 & 0.76 & \underline{1.60} & \textbf{4.50} & \textbf{1.26} \\
MonoCon~\cite{MonoCon}   & 0.40 & 1.08 & 1.78 & 0.85 & 0.78 & 1.65 & 6.02 & 1.40 \\
MonoUNI~\cite{MonoUNI}   & 0.38 & \underline{0.92} & 1.79 & 0.87 & \underline{0.72} & 1.79 & 4.98 & 1.43 \\
MonoCD~\cite{MonoCD}     & \underline{0.37} & 1.04 & \underline{1.72} & \underline{0.83} & 0.73 & \textbf{1.59} & 5.78 & 1.33 \\
MonoLaDN (ours)          & \textbf{0.32} & \textbf{0.87} & \textbf{1.54} & \textbf{0.81} & \textbf{0.55} & 1.62 & \underline{4.76} & \underline{1.28} \\
\bottomrule
\end{tabular}%
}
\label{tab:depth_mae}
\end{subtable}
\hfill
\begin{subtable}{0.48\textwidth}
\centering
\caption{Computational complexity.}
\setlength{\tabcolsep}{5pt}
\renewcommand{\arraystretch}{1.15}
\resizebox{\linewidth}{!}{%
\begin{tabular}{l|cccc}
\toprule
\multirow{2}{*}{Method} & Params & FLOPs $\downarrow$ & AP$_{3D|R40}$ $\uparrow$ & Runtime $\downarrow$ \\
 & (M) & (G) & (Moderate) & (ms) \\
\midrule
MonoDGP~\cite{MonoDGP}   & \textbf{42.16} & \textbf{68.99} & 18.72 & \textbf{42.4} \\
MonoCLUE~\cite{MonoCLUE} & 44.17 & 72.71 & 19.70 & 52.5 \\
MonoDLGD~\cite{MonoDLGD} & \textbf{42.16} & \underline{69.21} & \underline{19.87} & \underline{42.8} \\
\rowcolor{white}
MonoLaDN (ours) & \textbf{42.16} & \textbf{68.99} & \textbf{20.45} & \textbf{42.4} \\
\bottomrule
\end{tabular}%
}
\label{tab:complexity}
\end{subtable}
\end{table*}

\noindent\textbf{Cross-Dataset Evaluation.} To assess generalization, we follow DEVIANT~\cite{DEVIANT} and evaluate the KITTI-trained model on the car category of the nuScenes frontal validation set with zero fine-tuning. As shown in \cref{tab:depth_mae}, MonoLaDN attains the lowest depth MAE across all ranges on KITTI, and the lowest or second-lowest error on nuScenes despite the large domain shift. This indicates that the transferred geometric reasoning improves robustness to domain shift rather than overfitting to the source dataset. DEVIANT~\cite{DEVIANT} occasionally yields lower far-range error, owing to its equivariance to depth translations.

\noindent\textbf{Efficiency.} As shown in \cref{tab:complexity}, our proposed MonoLaDN maintains the same computational budget as its baseline MonoDGP~\cite{MonoDGP} while achieving a substantial boost in $AP_{3D}$. This is because the VLM and the entire language-guided reasoning branch are used only during training and fully discarded at inference, leaving the deployed detector identical to the baseline in parameters, FLOPs, and runtime. Furthermore, compared to the strongest competing methods MonoCLUE~\cite{MonoCLUE} and MonoDLGD~\cite{MonoDLGD}, MonoLaDN demonstrates clear dominance, outperforming both in 3D detection accuracy and inference efficiency.
\subsection{Ablation Study}
\label{sec:experiments:ablation_study}
We verify the effectiveness of each of our components in \cref{tab:main_ablation} (more ablation can be found in the supplementary material), reporting $\mathbf{AP_{3D}}$ for the \textit{Car} category on the KITTI \emph{val} set.

\noindent\textbf{Effectiveness of the reasoning branch.}
Starting from the MonoDGP~\cite{MonoDGP} baseline (a), we introduce the reasoning branch with the embedding loss $\mathcal{L}_{\text{emb}}$ (b). Since the perturbed queries are derived from the ground-truth boxes, the leakage-prevention mask $A^{lp}$ is enabled alongside the reasoning branch, preventing the learnable queries from accessing ground-truth reasoning as a shortcut, and is kept on in all subsequent settings. Moderate $AP_{3D}$ improves from 22.34 to 22.55, with consistent gains on easy and hard, showing that reconstructing the VLM reasoning embeddings from their noised form is a useful training signal for the detector's 3D geometry.

\noindent\textbf{Choice of text encoder.}
We next compare encoders for embedding the reasoning text. Replacing CLIP (ViT-B/32)~\cite{CLIP} (b) with SigLIP (base-patch16-224)~\cite{SigLIP} (c) raises moderate $AP_{3D}$ from 22.55 to 22.72 and improves both other settings, indicating that SigLIP yields representations better aligned with our geometric reasoning. We therefore adopt SigLIP in all remaining experiments.

\noindent\textbf{Effectiveness of the reconstruction objectives.}
We then study the two reconstruction losses in isolation. The object loss $\mathcal{L}_{\text{obj}}$ (d) alone reaches 23.15 moderate $AP_{3D}$, outperforming the embedding loss $\mathcal{L}_{\text{emb}}$ (c) at 22.72 and showing that reconstructing object-level targets is the stronger supervisory signal. Combining both (e) further improves the result to 23.61, confirming that the embedding and object objectives are complementary rather than redundant.

\noindent\textbf{Effectiveness of the Separated Reasoning mask.} Adding the Separated Reasoning mask $A^{sr}$ (f) raises moderate $AP_{3D}$ to 24.52. By routing each reasoning query to its related context, the reasoning queries of the same object and those of the same attribute across objects, $A^{sr}$ lets each attribute be reconstructed from informative rather than unrelated cues, improving fidelity.

\noindent\textbf{Choice of reasoning VLM.} Finally, we vary the model that generates the reasoning text. Replacing Qwen2.5-VL-7B (f) with InternVL3.5-8B (g) improves moderate $AP_{3D}$ from 24.52 to 24.83, with consistent gains on easy and hard, indicating that stronger VLM reasoning translates into more accurate detection. Using SpaceThinker-Qwen2.5-VL-3B (h), a variant explicitly fine-tuned for 3D spatial reasoning, gives the best result at 35.21, 25.03, and 21.60 $AP_{3D}$, despite being the smallest of the three models. We therefore adopt it as our reasoning VLM. This trend suggests that our method will benefit directly from future VLMs with stronger 3D reasoning, gaining accuracy without any change to the detector. Across all variants, the detector remains VLM-free at inference, so improved reasoning quality comes at no deployment cost.
\begin{table}[t]
\centering
\caption{\textbf{Ablation of each component on the KITTI \emph{val} set.} $\mathcal{L}_{\text{emb}}$/$\mathcal{L}_{\text{obj}}$ are the embedding and box reconstruction losses. $A^{lp}$/$A^{sr}$ the leakage-prevention and separated-reasoning masks. Text Encoder embeds the reasoning text, and VLM is the model that generates it. The best results are in \textbf{bold.}}
\label{tab:main_ablation}
\setlength{\tabcolsep}{4pt}
\renewcommand{\arraystretch}{1.05}
\resizebox{\textwidth}{!}{%
\begin{tabular}{c|cccccc|ccc}
\toprule
 & \multicolumn{6}{c|}{Ablation} & \multicolumn{3}{c}{Val, $AP_{3D|R40}$} \\
 & $\mathcal{L}_{\text{emb}}$ & $\mathcal{L}_{\text{obj}}$ & $A^{lp}$ & $A^{sr}$ & Text Encoder & VLM & Easy & Mod. & Hard \\
\midrule
(a) & $\times$    & $\times$    & $\times$    & $\times$    & $\times$             & $\times$            & 30.76 & 22.34 & 19.02 \\
(b) & \checkmark & $\times$    & \checkmark & $\times$    & CLIP~\cite{CLIP}     & Qwen2.5-VL~\cite{Qwen-VL}     & 31.41 & 22.55 & 19.34 \\
(c) & \checkmark & $\times$    & \checkmark & $\times$    & SigLIP~\cite{SigLIP} & Qwen2.5-VL~\cite{Qwen-VL}     & 31.63 & 22.72 & 19.53 \\
(d) & $\times$    & \checkmark & \checkmark & $\times$    & SigLIP~\cite{SigLIP} & Qwen2.5-VL~\cite{Qwen-VL}     & 32.07 & 23.15 & 19.98 \\
(e) & \checkmark & \checkmark & \checkmark & $\times$    & SigLIP~\cite{SigLIP} & Qwen2.5-VL~\cite{Qwen-VL}     & 33.46 & 23.61 & 20.57 \\
(f) & \checkmark & \checkmark & \checkmark & \checkmark & SigLIP~\cite{SigLIP} & Qwen2.5-VL~\cite{Qwen-VL}     & 34.78 & 24.52 & 21.13 \\
(g) & \checkmark & \checkmark & \checkmark & \checkmark & SigLIP~\cite{SigLIP} & InternVL3.5~\cite{InternVL} & 35.04 & 24.83 & 21.47 \\
(h) & \checkmark & \checkmark & \checkmark & \checkmark & SigLIP~\cite{SigLIP} & SpaceThinker~\cite{SpaceThinker} & \textbf{35.21} & \textbf{25.03} & \textbf{21.60} \\
\bottomrule
\end{tabular}%
}
\end{table}

\subsection{Qualitative Results}
\label{sec:experiments:qualitative_results}
\begin{figure*}[h!]
    \centering
    \includegraphics[width=0.88\textwidth]{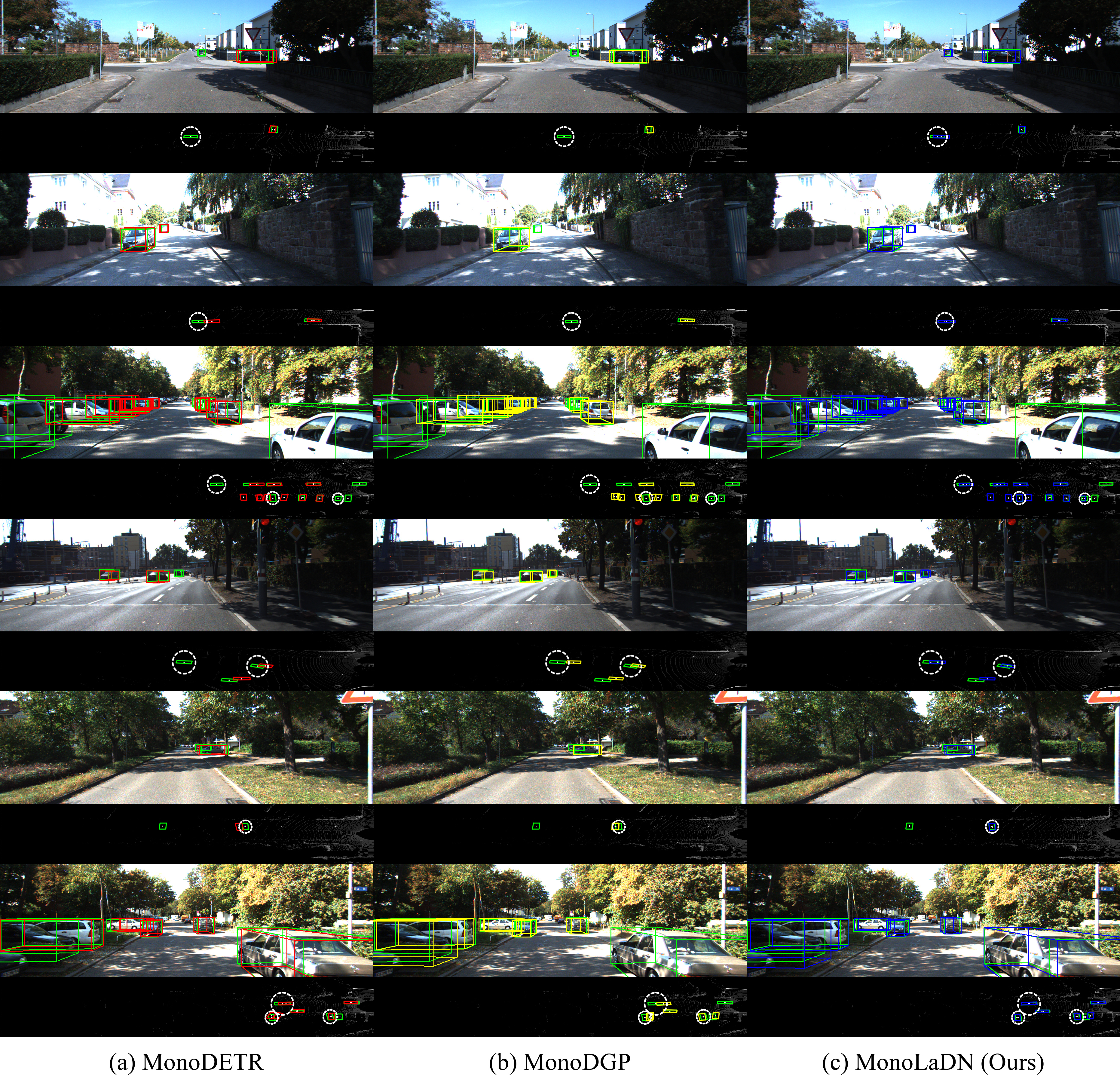}
    \caption{\textbf{Qualitative results on KITTI \emph{val} set.} (a) MonoDETR~\cite{MonoDETR}. (b) MonoDGP~\cite{MonoDGP}. (c) MonoLaDN (Ours). For each image set, the top row presents the camera-view visualization, while the bottom row offers the corresponding bird's-eye view. Ground-truth bounding boxes are rendered in \textcolor[rgb]{0,0.7,0}{green}, and predictions are shown in order: \textcolor{red}{red}, \textcolor[rgb]{0.9,0.7,0}{yellow}, \textcolor{blue}{blue}. We also circle some objects to highlight the difference between other state-of-the-art methods and our method.}
    \label{fig:qualitative}
\end{figure*}

\cref{fig:qualitative} presents qualitative comparisons on the KITTI \emph{val} set against MonoDETR~\cite{MonoDETR} and MonoDGP~\cite{MonoDGP}. The 3D boxes predicted by MonoLaDN align more closely with the ground truth in both the image and bird's-eye views, particularly in localization along the depth axis. The improvement is most evident for distant and partially occluded objects, where the baselines tend to misestimate depth or object extent, producing boxes that are shifted or poorly oriented in the bird's-eye view, whereas our predictions stay better aligned. These results are consistent with our quantitative gains and indicate that the geometric reasoning transferred from the VLM during training helps the detector recover more accurate 3D geometry.
\section{Conclusion}
\label{sec:conclusion}
In this work, we presented MonoLaDN, a framework that transfers geometric reasoning from a vision-language model into a monocular 3D detector while keeping inference purely vision-based. During training, a frozen VLM provides structured reasoning about occlusion, depth, dimension, and angle, which is encoded into noisy queries for reconstruction. A language-guided attention mask prevents ground-truth shortcuts and restricts each attribute to relevant context. The reasoning branch is removed at inference, introducing no additional cost. Experiments on KITTI 3D and nuScenes show consistent improvements over state-of-the-art detectors and strong cross-domain generalization.

\noindent\textbf{Limitations and future work.} Our approach currently targets single-image detection. Extending the Language-Guided DeNoising to multi-view and temporal settings is left for future work.

\section*{Acknowledgement} We acknowledge Ho Chi Minh City University of Technology (HCMUT), VNU-HCM for supporting this study.

\bibliographystyle{splncs04}
\bibliography{main}

\titlerunning{MonoLaDN}

\author{Kiet Dang Vu\inst{1,2}\orcidlink{0009-0004-0964-8134} \and
Cao Le Minh Khoa\inst{1,2}\orcidlink{0009-0008-5056-6578} \and
Nguyen Viet Tri Pham\inst{1,2}\orcidlink{0009-0002-4205-0148} \and
An Nguyen Gia\inst{1,2}\orcidlink{0009-0004-7743-7013}\and
Tam Le Thuc Ho\inst{1,2}\orcidlink{0009-0008-5445-280X}\and
Duc Dung Nguyen\inst{1,2}\thanks{Corresponding author: \href{mailto:nddung@hcmut.edu.vn}{nddung@hcmut.edu.vn}}\orcidlink{0000-0001-7321-7401}}

\authorrunning{Kiet Dang Vu et al.}

\institute{AITech Lab., Ho Chi Minh City University of Technology (HCMUT) \and Vietnam National University Ho Chi Minh City (VNUHCM)}

\clearpage
\setcounter{page}{1}
\appendix
\maketitlesupplementary

\begin{figure*}[h!]
    \centering
    \includegraphics[width=\textwidth]{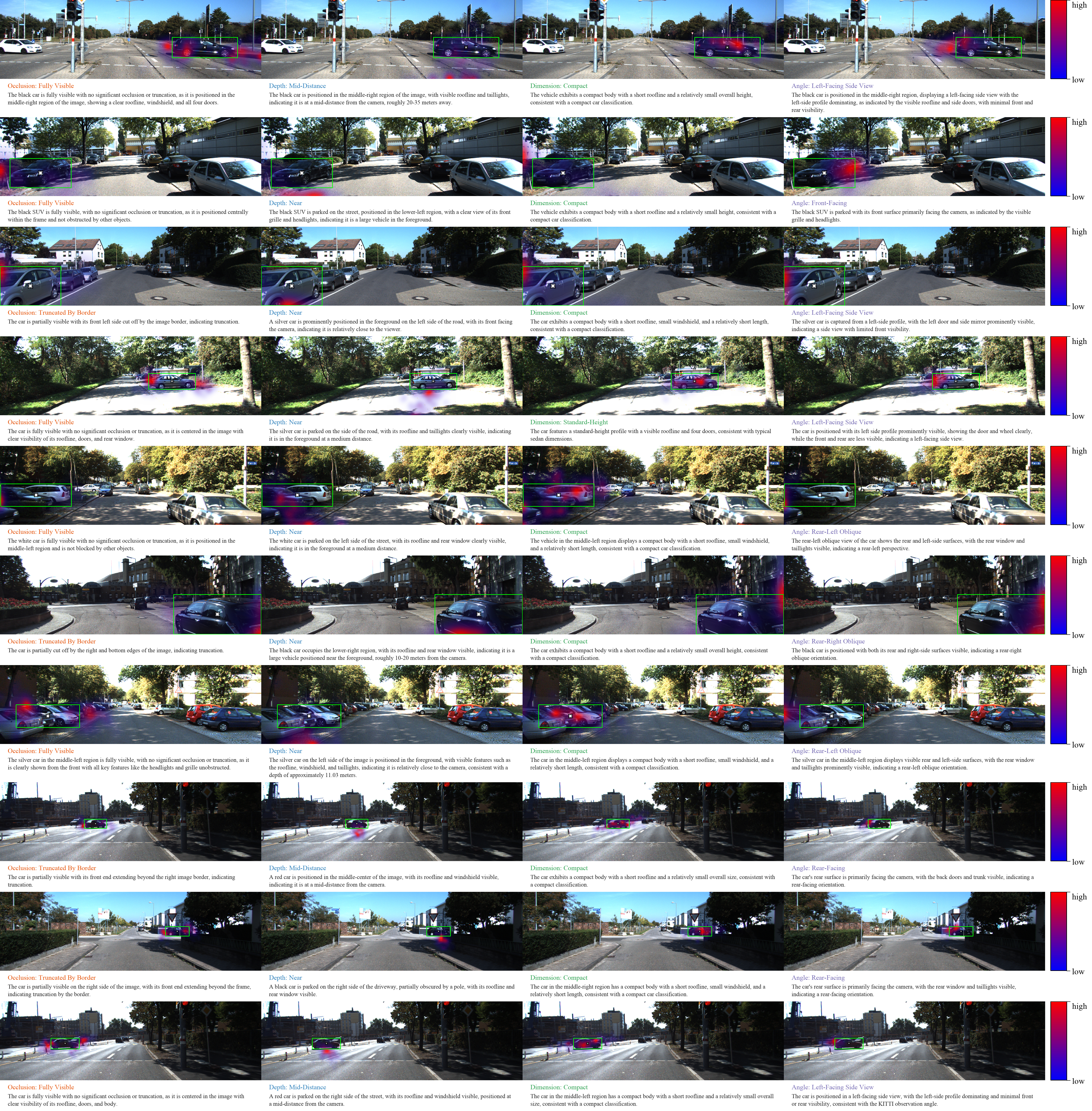}
    \caption{\textbf{Deformable cross-attention of the reasoning queries on the KITTI \emph{val} set.} Each row is one object, and the four columns show the occlusion, depth, dimension, and angle reasoning queries. For each query, we accumulate its deformable sampling locations at the final decoder layer, weighted by the corresponding attention weights, into a Gaussian heatmap overlaid on the image (red high, blue low). The target object is marked in \textcolor[rgb]{0,0.7,0}{green}, and the generated reasoning for each attribute is shown beneath its column.}
    \label{fig:supp:deformable_attention}
\end{figure*}
\section{Addition Explanation of MonoLaDN}
\label{sup:sec:explainMonoLaDN}
\subsection{Where Do the Reasoning Queries Attend?}
\label{sec:supp:deformable_attention}
To understand why the transferred reasoning improves detection, we visualize where the reasoning queries attend in the image. For each ground-truth object, we extract the deformable sampling locations and attention weights of its four attribute queries at the final decoder layer, and accumulate them into per-attribute Gaussian heatmaps over the input image (\cref{fig:supp:deformable_attention}).

The four attributes attend to clearly distinct and semantically meaningful regions, rather than collapsing onto a single object blob. The occlusion query spreads over the object boundary and the surrounding context, where evidence of truncation or overlap with neighboring objects appears; for truncated cases it concentrates toward the image border that cuts the object. The depth query attends to the ground-contact region and the scene context around the object, which carry the perspective cues needed to localize distance. The dimension query covers the full extent of the object body, consistent with reasoning about overall shape and size. The angle query focuses on orientation-revealing parts such as the visible side profile, front, or rear faces, matching the viewpoint stated in its reasoning.

Crucially, these attention patterns are consistent with the natural-language reasoning generated for each attribute. For instance, when the reasoning describes a left-facing side view, the angle query concentrates on the side profile of the vehicle, and when it reports truncation by the border, the occlusion query attends to the cut-off region. This alignment shows that the reasoning queries are spatially grounded in the image evidence rather than memorizing the ground-truth annotation, so reconstructing them forces the detector's depth-guided features to encode genuine, attribute-specific geometric cues that remain useful after the reasoning path is removed at inference.
\subsection{How Reasoning Queries Exchange Information?}
\label{sec:supp:self_attention}

\begin{figure*}[h!]
    \centering
    \includegraphics[width=\textwidth]{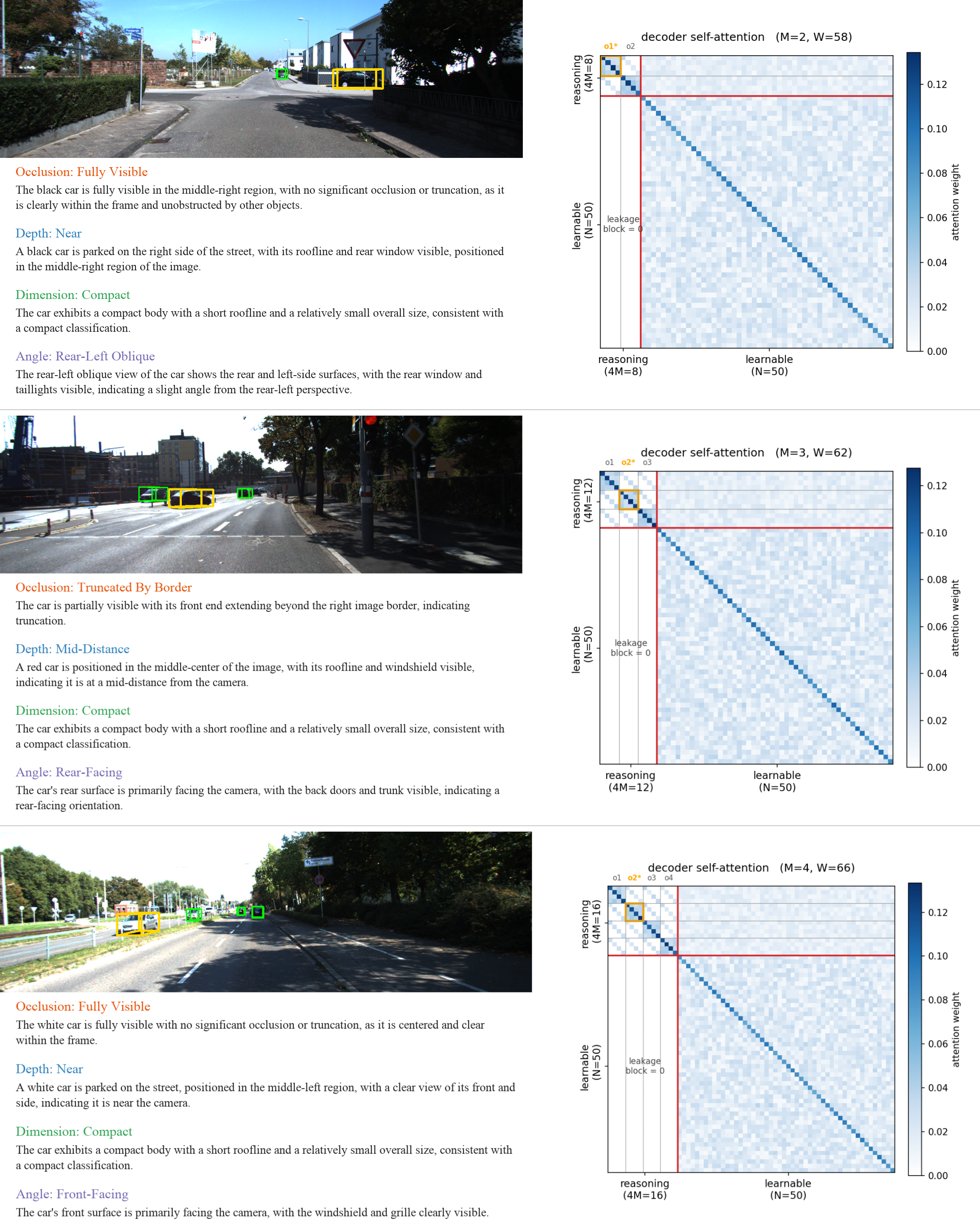}
    \caption{\textbf{Decoder self-attention under the Language-Guided Mask on the KITTI \emph{val} set.} Each row is one scene. The left column shows the image with all projected ground-truth 3D boxes (the target object in \textcolor[rgb]{0.9,0.6,0}{orange}, others in \textcolor[rgb]{0,0.7,0}{green}) and the generated reasoning for its four attributes. The right column shows the decoder self-attention matrix over the $W = 4M + N$ queries, ordered as the $4M$ reasoning queries followed by the $N=50$ learnable queries; the red lines separate the two groups and the thin gray lines delimit the four-attribute block of each object.}
    \label{fig:supp:self_attention}
\end{figure*}

We next examine how information flows among the queries inside the decoder, by visualizing the self-attention matrix together with the structure imposed by the Language-Guided Mask (\cref{fig:supp:self_attention}). The queries are ordered as the $4M$ reasoning queries followed by the $N=50$ learnable queries, matching the mask defined.

The learned attention respects the leakage-prevention mask exactly. The block formed by the learnable-query rows and the reasoning-query columns is empty, confirming that no learnable detection query attends to any reasoning query. The learnable queries therefore cannot copy the ground-truth-derived reasoning as a shortcut, which is essential since this path is removed at inference; without it, the detector would rely on information that is unavailable at test time and collapse, as shown in our mask ablation.

Within the reasoning block, the attention follows the separated-reasoning structure rather than spreading uniformly. Each object forms a dense $4\times4$ block on the diagonal, showing that the four attributes of the same object exchange information with one another, while a regular off-diagonal pattern links queries of the same attribute across different objects. This confirms that intra-object cross-attribute communication carries real weight, which is precisely why our separated-reasoning mask outperforms the stricter attribute-isolated variant that severs these links. Meanwhile, the reasoning queries attend to the learnable queries, so the geometric reasoning is injected into the detection queries during training. Together, these patterns show that the Language-Guided Mask routes information as designed: it blocks ground-truth shortcuts while letting each attribute reconstruct itself from informative context, allowing the detector's features to absorb the VLM's geometric reasoning that remains available after the reasoning path is discarded.
\section{Detail of 3D Reasoning Generation}
\label{sup:sec:3d_reasoning_gen}

This section details the offline reasoning-generation pipeline used to supervise the denoising branch. For each ground-truth object, we (i) derive a discrete ordinal label for each of the four attributes directly from the 3D annotation, (ii) assemble a prompt from these labels and object metadata, and (iii) query a frozen VLM to produce one evidence sentence per attribute. The entire process is run once before training and cached.

\subsection{Geometry-to-Language Labels}
\label{sup:sec:labels}
Rather than asking the VLM to regress continuous geometry, we map each annotation to a discrete label using fixed, deterministic rules. \cref{sup:tab:label_rules} summarizes the rules for occlusion, depth, and angle, and the dimension rules are described below.

\begin{table}[h]
\centering
\caption{\textbf{Geometry-to-language label rules.} Each attribute is mapped from the ground-truth annotation to a discrete ordinal label. $\alpha$ is the observation angle, $z$ the camera-frame depth, and $h/w/l$ the box height/width/length.}
\label{sup:tab:label_rules}
\setlength{\tabcolsep}{6pt}
\renewcommand{\arraystretch}{1}
\begin{tabular}{l|l|l}
\toprule
Attribute & Rule & Label \\
\midrule
\multirow{4}{*}{Occlusion}
 & trunc $\leq 0.15$, occ $= 0$   & Fully visible \\
 & occ $> 0$                       & Partially occluded \\
 & trunc $> 0.15$                  & Truncated by border \\
 & trunc $> 0.15$, occ $> 0$       & Truncated and occluded \\
\midrule
\multirow{5}{*}{Depth ($z$)}
 & $z \leq 10$      & Very near \\
 & $10 < z \leq 20$ & Near \\
 & $20 < z \leq 35$ & Mid-distance \\
 & $35 < z \leq 50$ & Far \\
 & $z > 50$         & Very far \\
\midrule
\multirow{4}{*}{Angle ($|\alpha|$)}
 & $|\alpha| \leq 25^\circ$          & Rear-facing \\
 & $|\alpha| \geq 155^\circ$         & Front-facing \\
 & $65^\circ \leq |\alpha| \leq 115^\circ$ & Left / right side view \\
 & otherwise                          & Oblique views \\
\midrule
\multirow{5}{*}{Dimension}
 & $h \geq 2.2$          & Tall \\
 & $w \geq 2.2$          & Wide \\
 & $l/h \geq 3.6$        & Elongated \\
 & $l \leq 3.2$          & Compact \\
 & otherwise             & Moderate-length \\
\bottomrule
\end{tabular}
\end{table}

\noindent\textbf{Occlusion.} The KITTI truncation and occlusion flags are thresholded (truncation $> 0.15$, occlusion $> 0$) and combined into one of four visibility labels, capturing whether the object is fully visible, occluded by another object, truncated by the image border, or both.

\noindent\textbf{Depth.} The camera-frame depth $z$ is binned by the boundaries $[10, 20, 35, 50]$~m into five distance labels, from \emph{very near} to \emph{very far}.

\noindent\textbf{Angle.} The observation angle $\alpha$ is mapped by its magnitude $|\alpha|$ into cardinal views (front/rear/left/right) and oblique views, with the sign of $\alpha$ selecting the left or right variant, giving eight viewpoint labels in total.

\noindent\textbf{Dimension.} The dimension label is class-dependent. Pedestrians are labeled \emph{Upright} or \emph{Compact}, cyclists \emph{Elongated}, and large vehicles (van, truck, tram) \emph{Long} or \emph{Boxy}. For cars, the label is chosen from the height $h$, width $w$, and length $l$: $h \geq 2.2$~m gives \emph{Tall}, $w \geq 2.2$~m gives \emph{Wide}, $l/h \geq 3.6$ gives \emph{Elongated}, $l \leq 3.2$~m gives \emph{Compact}, and the remaining cars are labeled \emph{Moderate-length}. This grounds the shape descriptor in the true box proportions while remaining a single discrete token.

\subsection{Prompt Construction}
\label{sup:sec:prompt}
Each attribute is prompted independently. For a given attribute, the frozen VLM is first queried to select one label from that attribute's ontology, using a system prompt that fixes the task and restricts the reasoning to visible object cues, and a user prompt containing the object metadata (class, camera context, range, region, scale, and border contact) together with the candidate labels and their definitions. Conditioned on the selected label, a second prompt then asks the VLM to write one concise evidence sentence justifying that label from the visible cues. The system prompt and object metadata are shared across the four attributes, while the candidate labels, their definitions, and the resulting evidence differ per attribute. To keep the evidence grounded in the object rather than the input construction, generated sentences are filtered to remove references to prompt artifacts such as crops or overlays.

\subsection{Output Format}
\label{sup:sec:output}
For each attribute, the VLM returns a \texttt{\{label, evidence\}} pair, and the four pairs are concatenated into the per-object reasoning set $\{R_{occ}, R_{depth}, R_{dim}, R_{angle}\}$. Each $R_a$ therefore consists of a discrete ordinal label grounded in the annotation and a short natural-language justification grounded in the image, which is subsequently embedded and denoised as described in the main paper.
\section{More Results}
\label{sup:sec:more_results}

\subsection{Other Categories on KITTI}
\label{sec:supp:other_categories}
Beyond the car category reported in the main paper, we evaluate MonoLaDN on the more challenging pedestrian and cyclist categories of the KITTI test set, as shown in \cref{tab:ped_cyc}. These classes are substantially harder, since their smaller size and greater shape variation provide weaker geometric cues. Despite this, MonoLaDN attains the best results across all six metrics, surpassing the strongest competing methods MonoDLGD~\cite{MonoDLGD} and MonoCLUE~\cite{MonoCLUE}. The consistent gains on both categories indicate that the geometric reasoning transferred from the VLM benefits detection beyond the car category and generalizes across object types.
\begin{table}[h!]
\centering
\caption{\textbf{Comparison on the KITTI test set for the pedestrian and cyclist categories} at IoU = 0.5, reporting $AP_{3D|R40}$. \textbf{Best} in bold.}
\label{tab:ped_cyc}
\setlength{\tabcolsep}{6pt}
\renewcommand{\arraystretch}{1}
\begin{tabular}{l|ccc|ccc}
\toprule
\multirow{2}{*}{Methods} & \multicolumn{3}{c|}{Pedestrian} & \multicolumn{3}{c}{Cyclist} \\
 & Easy & Mod. & Hard & Easy & Mod. & Hard \\
\midrule
GUPNet~\cite{GUPNet}     & 14.72 & 9.53 & 7.87 & 4.18 & 2.65 & 2.09 \\
MonoCon~\cite{MonoCon}   & 13.10 & 8.41 & 6.94 & 2.80 & 1.92 & 1.55 \\
DEVIANT~\cite{DEVIANT}   & 13.43 & 8.65 & 7.69 & 5.05 & 3.13 & 2.59 \\
MonoDDE~\cite{MonoDDE}   & 11.13 & 7.32 & 6.67 & 5.94 & 3.78 & 3.33 \\
MonoDETR~\cite{MonoDETR} & 12.65 & 7.19 & 6.72 & 5.12 & 2.74 & 2.02 \\
MonoDGP~\cite{MonoDGP}   & 15.04 & 9.89 & 8.38 & 5.28 & 2.82 & 2.65 \\
MonoDLGD~\cite{MonoDLGD} & 15.99 & 10.44 & 8.82 & 6.62 & 4.39 & 3.35 \\
MonoCLUE~\cite{MonoCLUE} & 16.18 & 10.45 & 8.75 & 5.93 & 3.20 & 2.94 \\
\midrule
MonoLaDN (Ours)          & \textbf{16.57} & \textbf{10.81} & \textbf{9.03} & \textbf{6.64} & \textbf{4.45} & \textbf{3.42} \\
\bottomrule
\end{tabular}
\end{table}

\subsection{Training Complexity}
\label{sec:supp:training_complexity}
While MonoLaDN adds no cost at inference, the auxiliary reasoning path introduces a modest overhead during training. \cref{tab:training_complexity} compares our method against the MonoDGP~\cite{MonoDGP} baseline in terms of training time and peak memory usage. Adding the reasoning branch increases the training time from 4.68 to 5.23 minutes per epoch and the peak memory from 14.35 to 20.45 GB, while raising moderate $AP_{3D}$ from 22.34 to 25.03. This overhead is incurred only during training, and since the VLM reasoning text is generated offline beforehand, it is not repeated each epoch. At inference, the reasoning path is fully discarded, so the deployed detector retains the exact cost of the baseline.
\begin{table}[h!]
\centering
\caption{\textbf{Training Complexity.}}
\label{tab:training_complexity}
\setlength{\tabcolsep}{6pt}
\renewcommand{\arraystretch}{1.2}
\begin{tabular}{l|ccc|c|c}
\toprule
\multirow{2}{*}{Method} & \multicolumn{3}{c|}{Val, $AP_{3D|R40}$} & Training time & Memory \\
 & Easy & Mod. & Hard & (min/epoch) $\downarrow$ & Usage (GB) $\downarrow$ \\
\midrule
MonoDGP        & 30.76 & 22.34 & 19.02 & 4.68 & 14.35 \\
MonoDGP + Ours & 35.21 & 25.03 & 21.60 & 5.23 & 20.45 \\
\bottomrule
\end{tabular}
\end{table}

\subsection{3D Reasoning Generation Complexity}
\label{sec:supp:reasoning_complexity}
Since the reasoning text is generated by a frozen VLM, we report the offline cost of this stage for the three VLMs in \cref{tab:reasoning_complexity}, alongside their val $AP_{3D}$. This generation is performed once before training and cached, so it adds no cost at inference and is not repeated across epochs. SpaceThinker-3B attains the best accuracy at a memory footprint of 9.35~GB, the lowest of the three, as its smaller size offsets the longer generation time. InternVL3.5-8B is the fastest to generate but requires the most memory, while Qwen2.5-VL-7B sits in between. In all cases, the cost is incurred only once and entirely offline, so the choice of reasoning VLM affects preprocessing time and memory but never the deployed detector.

\begin{table}[h!]
\centering
\caption{\textbf{3D Reasoning Generation Complexity.} Comparison of the VLMs used to generate the reasoning text, reporting val $AP_{3D|R40}$ for the \textit{Car} category alongside the total generation time over the dataset and the memory cost. The reasoning is generated offline during training only, so it adds no inference cost.}
\label{tab:reasoning_complexity}
\setlength{\tabcolsep}{6pt}
\renewcommand{\arraystretch}{1.2}
\begin{tabular}{l|ccc|c|c}
\toprule
\multirow{2}{*}{VLM} & \multicolumn{3}{c|}{Val, $AP_{3D|R40}$} & Total time & Memory \\
 & Easy & Mod. & Hard & (h) $\downarrow$ & Usage (GB) $\downarrow$ \\
\midrule
Qwen2.5-VL-7B~\cite{Qwen-VL}        & 34.78 & 24.52 & 21.13 & 220.7 & 9.35 \\
InternVL3.5-8B~\cite{InternVL}      & 35.04 & 24.83 & 21.47 & 73.6 & 17.45 \\
SpaceThinker-3B~\cite{SpaceThinker} & 35.21 & 25.03 & 21.60 & 230.6 & 9.35 \\
\bottomrule
\end{tabular}
\end{table}
\section{Additional Ablation Study}
\label{sup:sec:addition_ablation}
\subsection{Ablation on the Language-Guided Mask Design}
\label{sec:supp:mask_design}

\begin{figure}[h!]
    \centering
    \includegraphics[width=\textwidth]{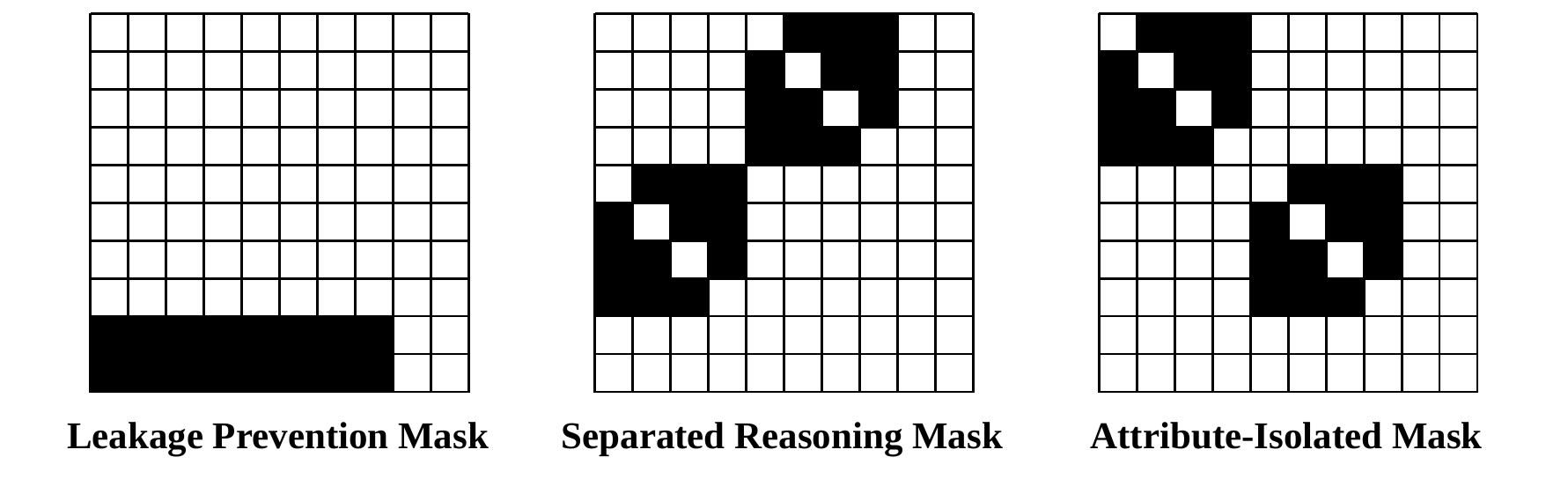}
    \caption{\textbf{Designs of the language-guided attention mask.} Black entries are blocked and white entries are permitted. The leakage-prevention mask $A^{lp}$ stops the learnable queries from attending to any reasoning query. The separated-reasoning mask $A^{sr}$ lets each reasoning query attend to all reasoning queries of the same object and to those of the same attribute across objects. The attribute-isolated mask $A^{ai}$ is a stricter variant that additionally blocks different attributes of the same object from attending to each other.}
    \label{fig:supp:mask_design}
\end{figure}

We study how the connectivity among reasoning queries affects denoising, comparing three mask rules of decreasing connectivity. Beyond the leakage-prevention mask $A^{lp}$ and our separated-reasoning mask $A^{sr}$, we introduce an attribute-isolated mask $A^{ai}$, a stricter rule that forbids different attributes of the same object from attending to one another. Denoting a reasoning query by its object index $o$ and attribute $a$, it is defined as
\begin{equation}
A^{ai}_{ij} =
\begin{cases}
0, & \text{if } i,j \in \mathbf{q}_P \text{ and } a_i \neq a_j, \\
1, & \text{otherwise,}
\end{cases}
\label{eq:attr_isolated_mask}
\end{equation}
so two reasoning queries are disconnected whenever they belong to different attributes, regardless of whether they share the same object. All experiments here use SigLIP~\cite{SigLIP} as the text encoder and Qwen2.5-VL-7B~\cite{Qwen-VL} as the reasoning VLM, varying only the mask. \cref{tab:supp:mask_ablation} reports $AP_{3D}$ for the \textit{Car} category on the KITTI \emph{val} set.

Removing $A^{lp}$ causes a severe collapse, dropping moderate $AP_{3D}$ from 23.61 to 10.92, since the learnable queries copy the ground-truth reasoning as a shortcut that is unavailable at inference~\cite{DN-DETR}. With $A^{lp}$ in place, adding our separated-reasoning mask $A^{sr}$ raises moderate $AP_{3D}$ from 23.61 to 24.52, confirming that routing each attribute to its informative context improves reconstruction. Replacing $A^{sr}$ with the stricter $A^{ai}$ instead yields only 23.81, and combining both further degrades performance to 23.42. These results show that exchanging information across the different attributes of the same object is beneficial, whereas isolating each attribute entirely is over-restrictive. Our separated-reasoning design therefore strikes the right balance and is adopted in all main experiments.

\begin{table}[t]
\centering
\caption{\textbf{Ablation on the language-guided mask design} on the KITTI \emph{val} set, reporting $AP_{3D|R40}$ for the \textit{Car} category. All rows use SigLIP as the text encoder and Qwen2.5-VL-7B as the reasoning VLM. $A^{lp}$, $A^{sr}$, and $A^{ai}$ denote the leakage-prevention, separated-reasoning, and attribute-isolated masks. The best results are in \textbf{bold}.}
\label{tab:supp:mask_ablation}
\setlength{\tabcolsep}{6pt}
\renewcommand{\arraystretch}{1.2}
\begin{tabular}{ccc|ccc}
\toprule
$A^{lp}$ & $A^{sr}$ & $A^{ai}$ & Easy & Mod. & Hard \\
\midrule
$\times$    & $\times$    & $\times$    & 17.25 & 10.92 & 9.87 \\
\checkmark & $\times$    & $\times$    & 33.46 & 23.61 & 20.57 \\
\rowcolor{yellow!40}
\checkmark & \checkmark & $\times$    & \textbf{34.78} & \textbf{24.52} & \textbf{21.13} \\
\checkmark & $\times$    & \checkmark & 33.64 & 23.81 & 20.72 \\
\checkmark & \checkmark & \checkmark & 33.05 & 23.42 & 20.11 \\
\bottomrule
\end{tabular}
\end{table}
\subsection{Ablation on the Noise Injection Strategy}
\label{sec:supp:noise_injection}

\begin{figure}[h!]
    \centering
    \includegraphics[width=\textwidth]{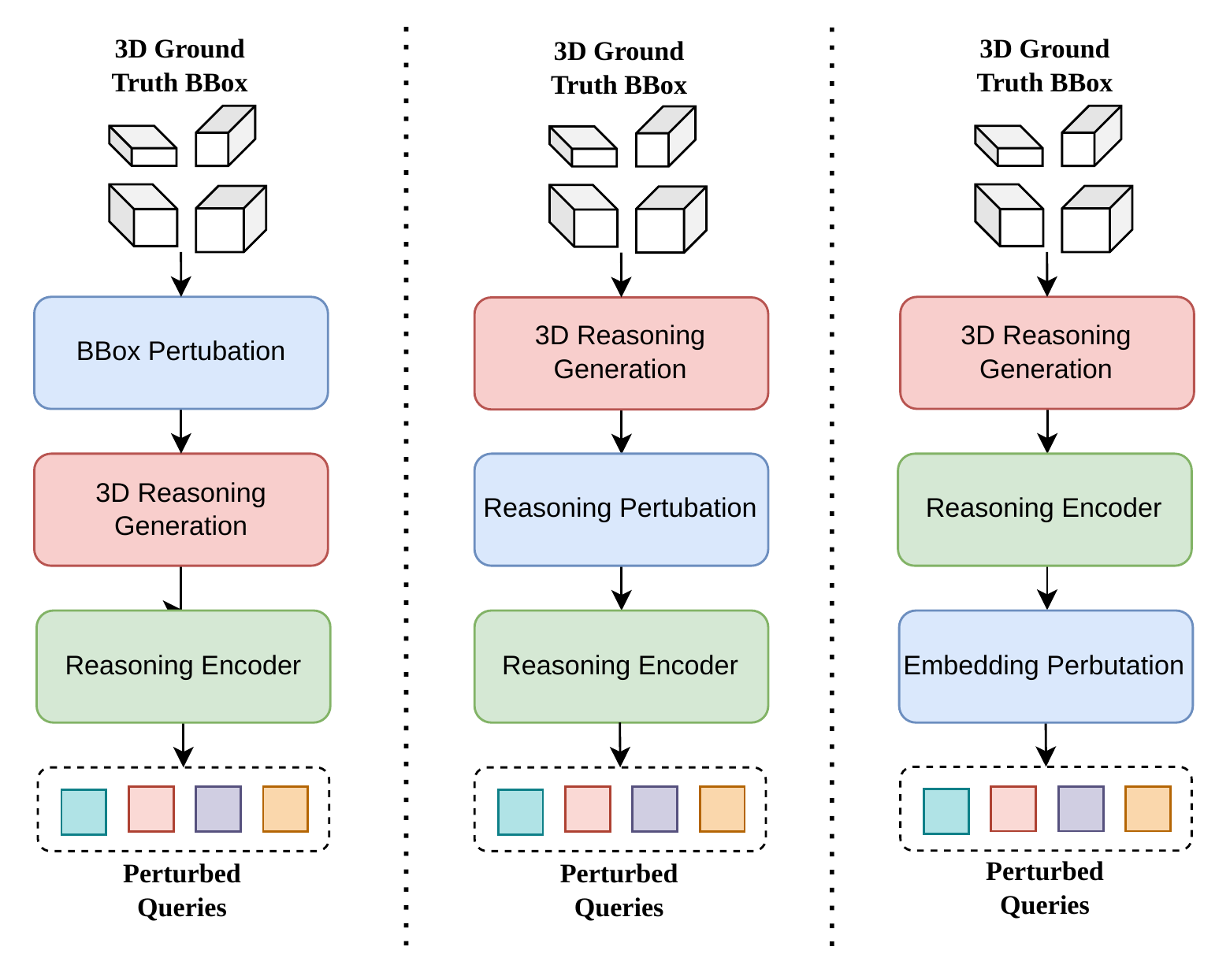}
    \caption{\textbf{Strategies for forming the perturbed queries.} All three start from the ground-truth 3D boxes and end with perturbed queries, but inject noise at different stages. \textbf{BBox perturbation} perturbs the boxes before the VLM, so the reasoning is generated from corrupted geometry. \textbf{Reasoning perturbation} generates clean reasoning and then applies text augmentation to the reasoning text before encoding. \textbf{Embedding perturbation} (ours) encodes the clean reasoning and injects learned noise in the embedding space.}
    \label{fig:supp:noise_injection}
\end{figure}

We compare three strategies for injecting the noise that forms the perturbed queries, differing in the stage at which corruption is applied, as illustrated in \cref{fig:supp:noise_injection}. \textbf{BBox perturbation} perturbs the ground-truth 3D boxes directly and then generates reasoning from the corrupted geometry. \textbf{Reasoning perturbation} generates reasoning from the clean boxes and applies text augmentation~\cite{textaug1,textaug2} to the reasoning text before encoding. \textbf{Embedding perturbation}, our main approach, encodes the clean reasoning with the text encoder and injects learned Gaussian noise in the embedding space via the $\beta$-VAE~\cite{VAE}. All experiments here use SigLIP~\cite{SigLIP} as the text encoder and Qwen2.5-VL-7B~\cite{Qwen-VL} as the reasoning VLM, varying only the noise strategy, and report $AP_{3D}$ for the \textit{Car} category on the KITTI \emph{val} set.

As shown in \cref{tab:supp:noise_ablation}, the stage at which noise is injected has a clear effect. BBox perturbation performs worst (22.83 moderate $AP_{3D}$): since the VLM is prompted with the perturbed boxes, the corrupted geometry no longer matches the visual evidence, so the VLM cannot reason reliably and produces inaccurate reasoning that provides a noisy, inconsistent supervision signal. Reasoning perturbation improves to 23.69 by keeping the geometry intact, but discrete text augmentation gives only coarse, non-differentiable control over the noise. Our embedding perturbation achieves the best result (24.52), since injecting continuous learned noise directly in the embedding space provides smooth, controllable corruption that is well matched to the reconstruction objective. We therefore adopt embedding perturbation in all main experiments.

\begin{table}[h!]
\centering
\caption{\textbf{Ablation on the noise injection strategy} on the KITTI \emph{val} set, reporting $AP_{3D|R40}$ for the \textit{Car} category. All rows use SigLIP as the text encoder and Qwen2.5-VL-7B as the reasoning VLM. The best results are in \textbf{bold}.}
\label{tab:supp:noise_ablation}
\setlength{\tabcolsep}{6pt}
\renewcommand{\arraystretch}{1.2}
\begin{tabular}{l|ccc}
\toprule
Noise Injection Strategy & Easy & Mod. & Hard \\
\midrule
BBox perturbation       & 33.15 & 22.83 & 19.65 \\
Reasoning perturbation  & 34.05 & 23.69 & 20.34 \\
\rowcolor{yellow!40}
Embedding perturbation (ours) & \textbf{34.78} & \textbf{24.52} & \textbf{21.13} \\
\bottomrule
\end{tabular}
\end{table}
\section{Additional Qualitative Results}
\label{sup:sec:addition_qualitative}
\begin{figure*}[h!]
    \centering
    \includegraphics[width=\textwidth]{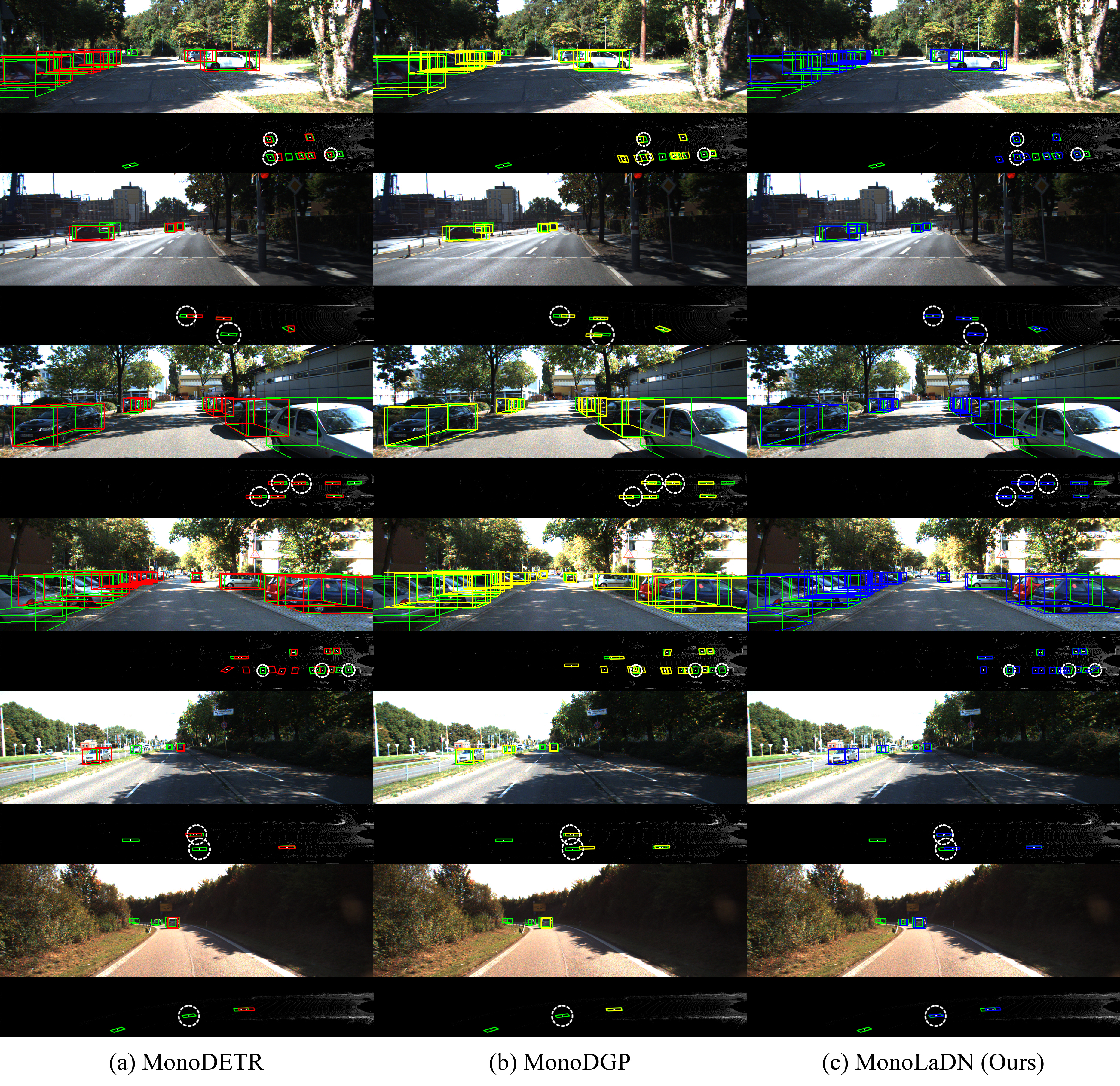}
    \caption{\textbf{Additional qualitative results on KITTI \emph{val} set.} (a) MonoDETR~\cite{MonoDETR}. (b) MonoDGP~\cite{MonoDGP}. (c) MonoLaDN (Ours). For each image set, the top row presents the camera-view visualization, while the bottom row offers the corresponding bird's-eye view. Ground-truth bounding boxes are rendered in \textcolor[rgb]{0,0.7,0}{green}, and predictions are shown in order: \textcolor{red}{red}, \textcolor[rgb]{0.9,0.7,0}{yellow}, \textcolor{blue}{blue}. We also circle some objects to highlight the difference between other state-of-the-art methods and our method.}
    \label{fig:supp_qualitative}
\end{figure*}

We provide further qualitative comparisons on the KITTI \emph{val} set in \cref{fig:supp_qualitative}, covering additional scenes with varying object density, distance, and occlusion. Consistent with the main paper, MonoLaDN produces 3D boxes that align more closely with the ground truth than MonoDETR~\cite{MonoDETR} and MonoDGP~\cite{MonoDGP}, particularly for distant and partially occluded objects.

\section*{Acknowledgement} We acknowledge Ho Chi Minh City University of Technology (HCMUT), VNU-HCM for supporting this study.
\end{document}